\documentclass{bmvc2k}

\title{LiDAR MOT-DETR: A LiDAR-based Two-Stage
Transformer for 3D Multiple Object Tracking}

\addauthor{Martha Teiko Teye}{m.teye-hk@uni-wuppertal.de}{1,2}
\addauthor{Ori Maoz}{ori.maoz@aptiv.com}{2}
\addauthor{Matthias Rottmann}{matthias.rottmann@uos.de}{3}

\addinstitution{
School of Mathematics and Natural Sciences\\
University of Wuppertal, Germany\\
}
\addinstitution{
 Aptiv Services Deutschland GmbH\\
 Wuppertal, Germany\\
}
\addinstitution{
Department of Mathematics, Computer Science and Physics\\
 Osnabr\"{u}ck University, Germany\\
}

\runninghead{Teye, Maoz, Rottmann}{LiDAR MOT-DETR}

\def\eg{\emph{e.g}\bmvaOneDot}
\def\Eg{\emph{E.g}\bmvaOneDot}
\def\etal{\emph{et al}\bmvaOneDot}

\usepackage{pifont}
\newcommand{\cmark}{\ding{51}}%
\newcommand{\xmark}{\ding{55}}%

\begin{document}

\maketitle

\begin{abstract}

Multi-object tracking from LiDAR point clouds presents unique challenges due to the sparse and irregular nature of the data, compounded by the need for temporal coherence across frames. Traditional tracking systems often rely on hand-crafted features and motion models, which can struggle to maintain consistent object identities in crowded or fast-moving scenes. We present a lidar-based two-staged DETR inspired transformer; a smoother and tracker. The smoother stage refines lidar object detections, from any off-the-shelf detector, across a moving temporal window. The tracker stage uses a DETR-based attention block to maintain tracks across time by associating tracked objects with the refined detections using the point cloud as context.
The model is trained on the nuScenes and KITTI datasets in both online and offline (forward peeking) modes demonstrating strong performance across metrics such as ID-switch and multiple object tracking accuracy (MOTA). The numerical results indicate that the online mode outperforms the lidar-only baseline and SOTA models on the nuScenes dataset, with an aMOTA of 0.724 and an aMOTP of 0.475, while the offline mode provides an additional $3\,$pp aMOTP. 

\end{abstract}

\section{Introduction}
\label{sec:intro}

Multi-object tracking (MOT) involves the detection and association of multiple objects across frames in an image or video sequence. MOT is very useful in applications such as automated driving, traffic control~\cite{trafficTrack}, video surveillance~\cite{videosurvillance} and animal behaviour studies~\cite{zhang2023animaltrack}.
The ability to accurately track multiple objects is a cornerstone of these systems, particularly in the context of self-driving vehicles and advanced robotics. LiDAR technology (Light Detection and Ranging) has become a key sensor in these domains due to its ability to provide precise 3D measurements of the environment~\cite{contreview2022,sun2020transtrack}. Numerous works have been conducted in 2D camera~\cite{motr_camera, TWB_2d, TasPoints2D, zhang2020fairmot}, radar domains~\cite{pan2024ratrack, zeller2024radar} and some extensions to 3D ~\cite{weng2020gnn3dmot, Mutr3d_camera}. 
However, MOT using LiDAR pointcloud (PC) data presents several challenges:  inaccurate motion models~\cite{yin2020lidar}, complex post-processing,   life cycle management and re-identification, short-term track identification of objects and inconsistent ego-motion of the sensor platform ~\cite{motmetrics-nus, wang2020pointtracknet}. For example, in the field of autonomous driving, most object tracking approaches require enormous (and tiring) optimization and hyperparameter tuning effort~\cite{zhang2020fairmot, mot20}, with the resulting tracker not being robust to changes in the data~\cite{weng2020gnn3dmot}.

\begin{figure*}
\begin{tabular}{cc}
\centering
\scalebox{0.5}{
\begin{tikzpicture}[
node distance=10mm, font=\small,
main_process/.style={rectangle, draw=red!60, very thick, short dashes, minimum size=3mm},
ffnnode/.style={rectangle, draw=black, fill=yellow!25, minimum size=10mm, text width =12mm, align=center},
featurenode/.style={rectangle, draw=black, fill=blue!25, minimum size=3mm, text width =13mm, align=center},
topknode/.style={rectangle, draw=cyan!60, fill=cyan!5, very thick, minimum size=3mm},
kindexnode/.style={rectangle, draw=cyan!60, fill=cyan!5, very thick, minimum size=3mm},
trackbox/.style={rectangle, draw=black, very thick, minimum size=1mm},
newtrackbox/.style={rectangle, draw=brown, very thick, minimum size=1mm},
encodernode/.style={trapezium, draw=black, fill=green!25, text width =7mm, align=center}
]

\node (detections1)[draw]{\includegraphics[width=0.1\textwidth]{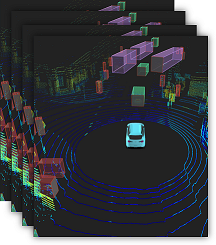}}; 
\node (det_text) [rectangle, below of=detections1, yshift=-5mm, xshift=-8mm] {$\mathcal{D}^{\tau-k, \ldots, \tau+k}$};
\node (tmpdet)[output, below of=detections1, yshift=0mm]{};
\node (fm2) [ffnnode, below of=tmpdet, yshift=-7mm] {Smoother};
\draw [arrow] (detections1) -- (fm2);

\node (tmpdet2)[output, below of=fm2, yshift=1mm]{};
\draw [arrow](fm2)--(tmpdet2);
\node (det1) [trackbox, below of=fm2,xshift=-5mm, yshift=0mm] {};
\node (det2) [trackbox, right of=det1,xshift=-7mm] {};
\node (det3) [trackbox, right of=det2,xshift=-7mm] {};
\node (det4) [trackbox, right of=det3,xshift=-7mm] {};
\node (det_text) [rectangle, right of=det4, yshift=1.5mm, xshift=-7.0mm] {$\mathcal{S}^{\tau}$};
\node (combine) [minimum size=1.5mm, right of=det4, xshift=0mm] {\(\oplus\)};

\draw [arrow] (det4) -- (combine);
\node (tracks_init) [process, fill=white!25, below of=tmpdet2, yshift=1mm, text width=9mm] {Tracks \\Init};
\node (qt_text) [rectangle, right of=tracks_init, yshift=1mm, xshift=-2.0mm] {$\mathcal{T}^{0}$};

\node (track01) [newtrackbox, right of=tracks_init, yshift=-1.5mm,xshift=-2.5mm, fill=red] {};
\node (track02) [newtrackbox, right of=track01,xshift=-7mm, fill=yellow] {};
\node (track03) [newtrackbox, right of=track02,xshift=-7mm, fill=blue] {};
\node (track04) [newtrackbox, right of=track03,xshift=-7mm, fill=green] {};
\draw [arrow] (tracks_init) -| (combine);

\node (pc_encoder) [featurenode, right of=fm2, xshift=15mm,] {Feature\\ Extraction};
\node (voxel) [rectangle, above of=pc_encoder, yshift=-1mm] 
{\includegraphics[width=0.04\textwidth]{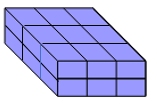}};
\node (pointclouds)[draw, above of=voxel, yshift=8mm]
{\includegraphics[width=0.1\textwidth]{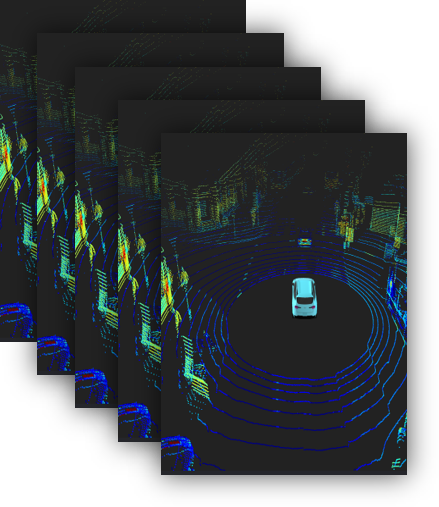}};
\node (dec) [encodernode, below of=pc_encoder, yshift=0mm] {\scriptsize DETR\\ decoder};
\draw [arrow] (combine) -- (dec);
\draw [arrow] (voxel) -= (pc_encoder);
\draw [arrow] (pc_encoder) -- (dec);
\draw[arrow] (pointclouds) --  (voxel);

\node (tracks) [process,  fill=white!25, below of=dec, yshift=0mm] {Tracks};

\draw [arrow] (dec) -- (tracks);

\node (track1) [trackbox, right of=tracks, yshift=1.5mm, xshift=-3mm, fill=red] {};
\node (track2) [trackbox, right of=track1,xshift=-7mm, fill=yellow] {};
\node (track3) [trackbox, right of=track2,xshift=-7mm, fill=blue] {};
\node (track4) [trackbox, right of=track3,xshift=-7mm, fill=green] {};

\node (qt_text1) [rectangle, right of=track4, xshift=-4.0mm] {$\mathcal{T}_{\tau}$};

\node (fm22) [ffnnode, right of=pc_encoder, xshift=18mm] {Smoother};
\node (tmpdet22)[output, below of=fm22, yshift=1mm]{};

\node (det11) [trackbox, below of=fm22,xshift=-5mm, yshift=0mm] {};
\node (det12) [trackbox, right of=det11,xshift=-7mm] {};
\node (det13) [trackbox, right of=det12,xshift=-7mm] {};
\node (det14) [trackbox, right of=det13,xshift=-7mm] {};
\node (det15) [trackbox, right of=det14,xshift=-7mm] {};
\node (det_text) [rectangle, right of=det15, yshift=1.5mm, xshift=-6.0mm] {$\mathcal{S}^{\tau+1}$};

\draw [arrow] (fm22) -- (tmpdet22);

\node (combine2) [minimum size=1.5mm, right of=det15, xshift=-3.7mm] {\(\oplus\)};
\draw [arrow] (tracks) -| (combine2);
\draw [arrow] (det15) -- (combine2);

\node (detections2)[draw, above of=fm22, yshift=17mm, xshift=0mm]{\includegraphics[width=0.1\textwidth]{images/stacked_det_comp.png}};

\draw [arrow] (detections2) -- (fm22);
\node (pc_encoder2) [featurenode, right of=fm22, xshift=15mm] {Feature\\extraction};
\node (voxel2) [rectangle, above of=pc_encoder2, yshift=-1mm] 
{\includegraphics[width=0.04\textwidth]{images/voxel2.png}};
\node (pointclouds2)[draw, above of=voxel2, yshift=8mm, xshift=0mm]{\includegraphics[width=0.1\textwidth]{images/stacked_pc.png}};
\node (dec2) [encodernode, below of=pc_encoder2, yshift=0mm] {\footnotesize DETR\\ decoder};
\node (tracks2) [process,  fill=white!25, below of=dec2, yshift=0mm] {Tracks};
\node (track11) [trackbox, right of=tracks2, xshift=-2mm, yshift=1.5mm, fill=red] {};
\node (track12) [trackbox, right of=track11,xshift=-7mm, fill=yellow] {};
\node (track13) [trackbox, right of=track12,xshift=-7mm, fill=blue] {};
\node (track14) [trackbox, right of=track13,xshift=-7mm, fill=green] {};
\node (track15) [newtrackbox, right of=track14,xshift=-7mm, fill=violet] {};

\node (tmptrack)[rectangle, right of=tracks2, xshift=18mm]{\ldots};
\draw[arrow] (tracks2) -- (tmptrack);

\node (qt_text2) [rectangle, below of=track12, yshift=5mm, xshift=1.0mm] {$\mathcal{T}^{\tau+1}$};

\node [output, left of=track2](tmp1){};

\draw [arrow] (voxel2) -= (pc_encoder2);
\draw [arrow] (pc_encoder2) -- (dec2);
\draw [arrow] (dec2) -= (tracks2);
\draw[arrow] (combine2) -- (dec2);
\draw[arrow] (pointclouds2) -- (voxel2);
\end{tikzpicture}} &

\scalebox{0.55}{
\begin{tikzpicture}[
node distance=1.15cm, font=\footnotesize,
mha/.style={rectangle, draw=black, fill=green!35, minimum size=3mm, text width=15mm},
mhac/.style={rectangle, draw=black, fill=green!35, minimum size=3mm, text width=16.5mm},
addnorm/.style={rectangle, draw=black,  fill=olive!45, minimum size=3mm},
main_process/.style={rectangle, draw=red!35, short dashes, minimum size=3mm},
ffnnode/.style={rectangle, draw=black, fill=brown!35, minimum size=3mm},
topknode/.style={rectangle, draw=cyan!60, fill=cyan!5, very thick, minimum size=3mm},
kindexnode/.style={rectangle, draw=cyan!60, fill=cyan!5, very thick, minimum size=3mm},
background rectangle/.style={fill=yellow!25}, show background rectangle,
]

\node (add_norm_enc) [addnorm] {Add \& Norm};

\node (multihead_enc) [mha, below of=add_norm_enc, yshift=3mm] {Multi-head\\attention};

\node (ffn1_enc) [ffnnode, above of=add_norm_enc, yshift=-4mm] {FFN};
\node (add_norm_enc2) [addnorm, above of=ffn1_enc, yshift=-4.5mm] {Add \& Norm};
\node (combine1_enc) [minimum size=0.5mm, below of=multihead_enc, yshift =3mm] {\(\oplus\)};
\node (rawdet) [output, below of=combine1_enc, yshift=3mm] {};
\node (rawdet2) [rectangle, right of=rawdet] {$\mathcal{D}^{\tau-k, \ldots, \tau+k}$};


\node (embedquery) [rectangle, right of=combine1_enc, xshift=3mm] {$\mathcal{P}^{\tau-k,\ldots,\tau+k}$};
\node (tmpdec) [output, left of=combine1_enc, xshift =-3mm, yshift =3.5mm] {};
\node (tmpdec2) [output, above of=combine1_enc, yshift =-7.5mm] {};

\node (tmpdec3) [output, above of=add_norm_enc,xshift =-0mm,  yshift =-7.5mm] {};
\node (tmpdec4) [output, left of=tmpdec3, xshift =-3mm] {};

\draw [arrow] (combine1_enc) -- (multihead_enc);
\draw [arrow] (multihead_enc) -- (add_norm_enc);
\draw [arrow] (add_norm_enc) -- (ffn1_enc);
\draw [arrow] (ffn1_enc) -- (add_norm_enc2);
\draw[arrow] (rawdet2) -| (combine1_enc);
\draw[arrow] (embedquery) -- (combine1_enc);
\draw[line] (tmpdec) -- (tmpdec2);
\draw[arrow] (tmpdec)|-(add_norm_enc);
\draw[line] (tmpdec3) -- (tmpdec4);
\draw[arrow] (tmpdec4)|-(add_norm_enc2);

\node (e1n_text) [rectangle, above of=add_norm_enc2, yshift=-4mm, xshift=0mm] {$\mathcal{E}^{\tau-k,\ldots,\tau+k}$};
\node [output, right of=e1n_text, xshift=3mm](tmpenc){};

\draw [arrow] (add_norm_enc2) -- (e1n_text);
\node (ffn1) [ffnnode, right of=e1n_text, xshift=13mm, yshift =-5mm] {FFN};

\node (sigmoid) [startstop, below of=ffn1, yshift =4mm,fill=white!35] {Sigmoid};
\node (topk) [process, below of=sigmoid, yshift=4mm] {Top-k};
\node [output, right of=topk, xshift=-5mm](tmp2){};

\node (index_z) [kindexnode, below of=topk] {Index};

\node [output, right of=index_z, xshift=-5mm](tmp3){};
\node (ffn3) [ffnnode, right of=index_z, yshift=2.5mm] {FFN};
\node (ffn4) [ffnnode, below of=ffn3, yshift=5mm] {FFN};

\node (oq) [rectangle, right of=ffn3,yshift=2mm, xshift=-3mm] {\(Q^\tau_{1,\ldots,L}\)};
\node (pe) [rectangle, right of=ffn4, yshift=-2mm, xshift=-3mm] {\(P^\tau_{1,\ldots,L}\)};

\node (tmpdec) [output, right of=ffn3, xshift =3mm] {};
\node (combine2) [minimum size=0.5mm, right of=ffn4, xshift =3mm] {\(\oplus\)};
\node (multihead) [mha, right of=pe, yshift=-3mm, xshift =14mm ] {Multi-head\\attention};
\node (add_norm) [addnorm, above of=multihead, yshift=-3.5mm] {Add \& Norm};
\node (tmp_add_norm) [output, left of=add_norm] {};
\node (tmpdec2) [output, above of=add_norm, yshift=-7.5mm,] {};
\node (tmpdec6) [output, right of=tmpdec2, xshift =-1mm] {};
\node (multiheadcross) [mhac, above of=add_norm, yshift=-3.5mm] {\scriptsize Multi-head \\cross-attention};
\node (add_norm2) [addnorm, above of=multiheadcross, yshift=-3.5mm] {Add \& Norm};
\node (e1n_text2) [rectangle, left of=add_norm2, xshift=-6mm] {$\mathcal{E}^{\tau-k,\ldots,\tau+k}$};
\node (e1n_text3) [output, left of=multiheadcross, xshift=-6mm] {};
\node (tmpdec3) [output, right of=add_norm2, xshift =-1mm] {};
\node (ffn5) [ffnnode, above of=add_norm2, yshift=-4mm] {FFN};
\node (tmpdec4) [output, above of=add_norm2, yshift =-7.5mm] {};
\node (tmpdec5) [output, right of=tmpdec4, xshift=-1mm] {};
\node (add_norm3) [addnorm, above of=ffn5, yshift=-5mm] {Add \& Norm};
\node (filtereddet) [rectangle, above of=add_norm3, yshift=-3mm] {\(\mathcal{S^\tau}\)};

\draw [line] (e1n_text) -| (tmpenc);

\draw [arrow] (tmpenc)  |- (ffn1);
\draw [arrow] (ffn1) -- (sigmoid);
\draw [arrow] (sigmoid) -- (topk);

\draw [arrow] (index_z) -- (tmp3) |- (ffn3);
\draw [arrow] (index_z) -- (tmp3) |- (ffn4);
\draw [arrow] (topk) --(index_z);
\draw [arrow] (rawdet2) -| (index_z);

\draw [arrow] (ffn4)  -- (combine2);
\draw [arrow] (combine2)  -| (tmp_add_norm)--(add_norm);
\draw [arrow] (e1n_text2)  |-(multiheadcross);
\draw [arrow] (ffn3)  -- (tmpdec)--(combine2);
 \draw [arrow] (combine2) |- (multihead);
\draw [arrow] (multihead) -- (add_norm);
\draw [arrow] (add_norm) -- (multiheadcross);
\draw [arrow] (multiheadcross) -- (add_norm2);
\draw [arrow] (add_norm2) -- (ffn5);
\draw [arrow] (ffn5) -- (add_norm3);
\draw [arrow] (tmpdec2)  -- (tmpdec6)|-(add_norm2);
\draw [arrow] (tmpdec4)  -- (tmpdec5)|-(add_norm3);
\draw [arrow] (add_norm3) -- (filtereddet);
\end{tikzpicture}
}
\\
(a) Overall architecture&(b) Smoother selection mechanism
\end{tabular}

\caption{LiDAR MOT-DETR works by first selecting relevant detections across a temporal window, k, using our Smoother mechanism. Relevant objects are then jointly tracked using detection queries, track queries and PC features in an iterative transformer track process. Brown background represents new tracks whiles existing tracks with black background.} 
\label{fig:overall}
\end{figure*}

Early (classical) tracking methods, which are still popular today, rely on explicit motion modelling as dynamical systems (for an introduction see e.g.\ \cite{simpletrack}). These approaches,  which includes Kalman Filter~\cite{deepsort} and particle filter~\cite{lee2022moving}, suffer from imperfect appearance and motion models~\cite{motionchallenge, depthchallenge} and are sensitive to the choice of hyper-parameters. They also strongly depend on ego-motion information (to compensate for induced motion in the sensors reference frame) that is often estimated separately and can be inefficient and error-prone~\cite{deepsort}. 

Neural networks have become a powerful tool for MOT due to their ability to learn robust feature representations~\cite{FastPoly, 3dMotformer, NEBP2022, VoxelNext2023}. Hybrid approaches that incorporate deep learning focused on modelling spatial and motion features inherently within the networks have improved the efficiency of object tracking~\cite{deepsort, NEBP2022, wang2023exploringobjectcentrictemporalmodeling}, though their performance largely depends on the performance of the underlying object detectors ~\cite{twodetector}. Transformers~\cite{vaswani2017attention}, originally developed for natural language processing, have proven remarkably versatile across various domains, including computer vision and, more recently, pointcloud processing~\cite{li3detr}.  Their ability to model long-range dependencies and handle sequences of varying lengths makes them well-suited for the task of MOT in LiDAR data~\cite{PointTransformer_2021_ICCV, wang2020pointtracknet}. \\
Our main contributions are as follows:
\begin{itemize}
\item We introduce LiDAR MOT-DETR, a modular, detector-agnostic (i.e.\ works with any existing object detector), two-stage framework which can be tailored for both online and offline tracking. 
\item Both smoother and tracker architectures are based on the DETR transformer, tailored specifically to the needs of smoothing and tracking. 
\item Our tracker outperforms current state-of-the-art (SOTA) lidar models with an aMOTA of 0.724 and aMOTP of 0.475 on nuScenes validation set in an online setting. Even more significantly, it achieves an  aMOTP of 0.445 on the nuScenes test set in offline mode.


\end{itemize}

\section{Related Work} \label{sec:work}
LiDAR-based Multiple Object Tracking has seen significant developments in recent years, particularly with the incorporation of deep learning methods~\cite{contreview2022, lidarreview, li3detr}:

\paragraph{Introduction of Multiple Object Tracking.}
Traditionally, MOT relied heavily on probabilistic models such as the Kalman Filter~\cite{simpletrack} and its variations, which use explicit motion modelling and heuristic track assignments such as greedy and Hungarian assignment~\cite{deepsort}. Kalman Filters enabled efficient data association, allowing for the matching of detected objects across frames. However, they often struggled with occlusions, dynamic environments, and complex object interactions. Deep learning methods, particularly Convolutional Neural Networks (CNNs), have transformed MOT by enabling robust feature extraction and learning from large datasets~\cite{FastPoly, 2023shasta} in both 2D and 3D. Approaches like FairMOT~\cite{zhang2020fairmot} and Deep SORT~\cite{deepsort} extended Kalman Filters  by combining CNN-based object detection features with Recurrent Neural Networks (RNNs) for temporal modelling, achieving state-of-the-art performance in image-based MOT. However, these methods are ineffective with LiDAR data due to its sparse and irregular nature~\cite{yin2020lidar}.

\paragraph{LiDAR-based Multiple Object Tracking.}
LiDAR sensors provide precise 3D point clouds that are invaluable for understanding the spatial arrangement of objects in an environment~\cite{contreview2022}. Early approaches to LiDAR-based MOT often relied on voxelization or point-wise feature extraction methods~\cite{bahraini2019slam}, followed by traditional tracking algorithms like the Extended Kalman Filter (EKF)~\cite{du2023strongsort, deepsort}. These methods were limited by their reliance on hand-crafted features and their inability to capture the complexity of object dynamics in 3D space. Recent advancements have seen the integration of deep learning models directly with LiDAR point clouds. Yin et al.~\cite{Centerpoint} introduced CenterPoint, a voxel-based framework that detects objects and tracks using greedy closest-point matching in 3D space. CenterPoint demonstrates that leveraging spatial features directly from point clouds significantly improves tracking accuracy. Utilizing further features from these detection models have provided recent deep learning trackers with enriched depth and spatial information which has been a baseline for most deep learning detectors~\cite{Qi2021Offboard3O} and trackers ~\cite{du2023strongsort, deepsort, 2023shasta}.

\paragraph{Transformers in LiDAR Multiple Object Tracking.}

In MOT, transformers have been utilized to enhance temporal consistency in tracking~\cite{3dMotformer, focalformer3d}. Wang et al.~\cite{Wang2024MCTrackAU} presented a strong sensor-fusion transformer-based framework for egomotion-aware 3D object tracking, which directly models the motion of objects and the sensor platform. 
Ruppel et al.~\cite{pctrack} developed a transformer-based tracker that builds on previous point-based detection systems~\cite{wang2020pointtracknet}. 
The inclusion of transformers enhanced the model’s ability to use rich query feature representations to track objects using attention-based mechanism~\cite{wang2023exploringobjectcentrictemporalmodeling, Mutr3d_camera, motr_camera, chen2023trajectoryformer3dobjecttracking} which is used in camera and lidar domain.
However, deep learning approaches often fall short with handling new-born trajectories~\cite{motr_camera}, mangaging bounding box precision, complex post-processing methods~\cite{motr_camera}, managing identity switches and the track life cycle~\cite{3dMotformer}. In this paper, we focus on inherently tracking objects using transformers without the need for heuristics track life cycle management.

\begin{figure}
\centering
\begin{tabular}{cc}

\includegraphics[width=3.8cm]{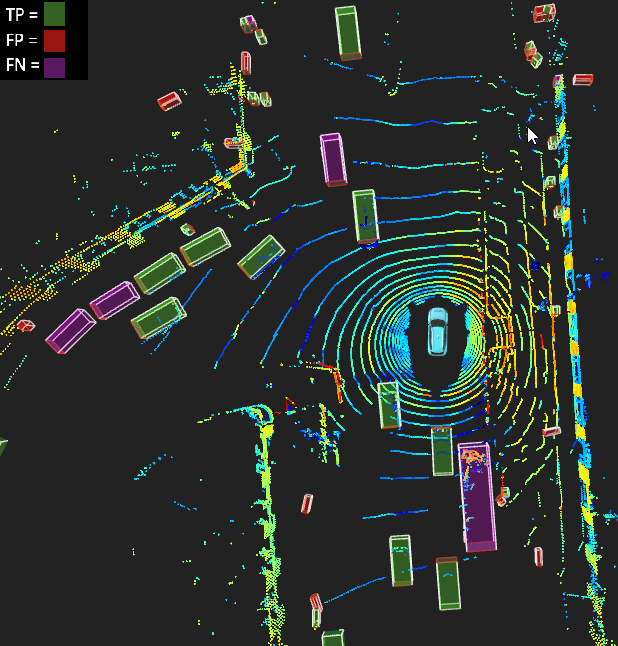} &
\includegraphics[width=3.8cm]{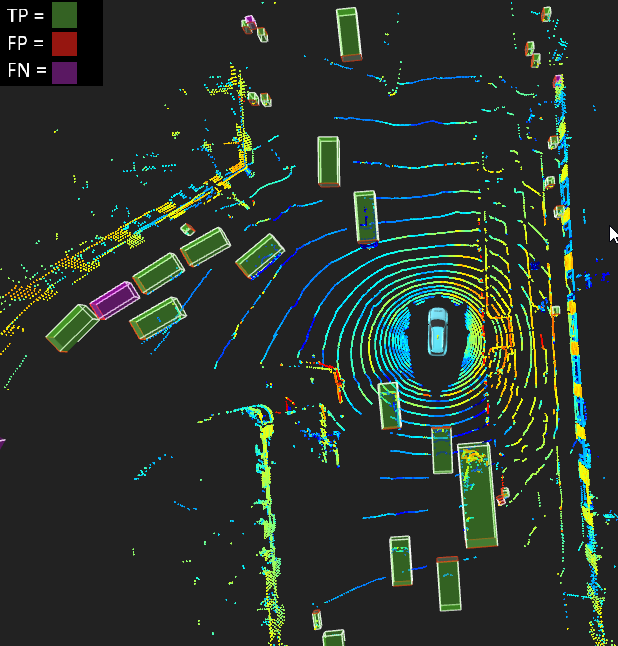}\\

  3DMOTFormer \cite{3dMotformer} & (Ours)

\end{tabular}
\caption{Qualitative comparison between our method and  another SOTA method on nuScenes validation set (scene\_token: 01452fbfbf4543af8acdfd3e8a1ee806). Both methods use the same base object detector (CenterPoint \cite{Centerpoint}). }
\label{fig:demo1}
\end{figure}
\section{Methodology} \label{sec_work}

We now introduce our novel two\hyp{}stage DETR~\cite{detrE2E} transformer-based architecture, consisting of a smoother and a tracker as depicted in figure \ref{fig:overall}. The smoother takes the bounding boxes from the object detector across multiple frames (either past \& future frames, or past only) and outputs cleaner and more consistent bounding boxes. The tracker takes the track predictions from the previous frames, the detected objects from the current frame and the raw point cloud, and outputs track predictions for the current frame.

\paragraph{Overview of Tracking and Smoothing.}
We start with a series of LiDAR point clouds and corresponding $\mathcal{D}^{\tau} = \{ d_1^{\tau}, \ldots, d_{n^{\tau}}^{\tau} \}$ sets of detected objects, from an existing (off-the-shelf) object detector, for each frame $\tau=1,\ldots,T$, with $n^{\tau}$ being the number of detections in frame $\tau$.  We assume that $n^{\tau}$ is bound by above by some $n \in \mathbb{N}$.

Each $d_{i}^{\tau}$ consists of a predicted class label, 3D bounding box center, extents and rotation angles.
For $\tau \notin \{1,\ldots,T\}$, let $D^{\tau} = \emptyset$. For some $k \in \mathbb{N}$, the smoother $f$ processes $t'-t+1$ frames $\mathcal{D}^{t, \ldots, t'} = (\mathcal{D}^{t},\ldots, \mathcal{D}^{t'})$, providing smoothed detections for frame $\tau$ as output, i.e., $\mathcal{S}^{\tau} = f(\mathcal{D}^{t, \ldots, t'}) $.
We set $t=\tau-k$ as well as $t'=\tau+k$, such that $\mathcal{D}^{\tau-k, \ldots, \tau+k}$ contains $2k+1$ frames with the key frame $\tau$ right at the centre of the sequence. 
The corresponding loss function $\ell_f$ for learning the smoothing is discussed below.

The tracker $g$ predicts the tracked objects in each frame based on the tracks of the previous frames, the detected objects from the current frame and the raw point cloud. It creates a set of track queries which includes the tracks from the previous frame $\mathcal{T}^{\tau-1}$ and newly-initialized tracks from any detection from  $\mathcal{S}^{\tau}$ which could not be matched to the existing tracks. It outputs bounding boxes $\mathcal{T}^{\tau}=\{p_{i}, s_{i}, \theta_{i} ,C_i, \mathit{tid}_{i} \}_{i=1,\ldots,N^\tau}$ consisting of the center position, extent, yaw rotation, score and track id, respectively, with  $N^\tau \leq \tau \cdot n$ denoting the total number of bounding boxes in frames $1,\ldots,\tau$.

\paragraph{Discussion of Loss Functions.} 

The smoother compares the smoothed detections $\mathcal{S}_{i}^{\tau}$ to the ground-truth $\mathcal{G}_{j}^{\tau}$, $i,j=1,\ldots,n$, while ignoring their ordering. It uses the same bipartite matching as proposed in DETR (DEtection TRansformers)~\cite{detrE2E}. Suppose we have $n_s$ smoothed detection and $n_g$ ground-truth boxes in frame $\tau$. Let $B \subseteq \{ (i,j) : i=1,\ldots,n_s, \; j=1,\ldots, n_g, \}$ be a bipartite matching such that for an elements $B_\iota$ and $B_{\iota'}$ with $\iota \neq \iota'$ with their two slots fulfilling $B_{\iota,\kappa} \neq B_{\iota',\kappa}$ for $\kappa=1,2$. We express the loss functions that we use as:
\begin{equation} \ell_f = \arg \min_B \sum_{(i,j) \in B} L_\mathit{match}(\mathcal{S}_{i}^{\tau}, \mathcal{G}_{j}^{\tau})\; \text{, ~~where  } L_{match}=W_\mathit{bbox}L_\mathit{bbox} + W_\mathit{cls} L_\mathit{cls}\; \label{eq:1}\end{equation} 
\(L_\mathit{bbox}\) denotes the L1 loss which can also be expressed as \(L_\mathit{bbox} = L_\mathit{box}(\mathcal{S}^{\tau}_i,\mathcal{G}^{\tau}_{j})\) and \(L_\mathit{cls}\) denotes the focal loss which also represents the association loss \cite{focalloss}. The loss weights are denoted by \(W_\mathit{bbox}\) and \(W_\mathit{cls}\), respectively.
The tracker uses the same loss as in smoother, with an additional \(L_\mathit{giou}\) \cite{giou} generalized intersection over union loss:
\begin{equation} \ell_g = \arg \min_B \sum_{(i,j) \in B} L'_\mathit{match}(\mathcal{T}_{i}^{\tau}, \mathcal{G}_{j}^{\tau}) \;\text{ , ~~where } L'_\mathit{match}=W'_\mathit{bbox}L_\mathit{bbox} + W'_\mathit{cls} L_\mathit{cls} +  L_\mathit{giou}\label{eq:2}\;\end{equation}
where the corresponding loss weights are denoted by \(W'_{bbox}\) and \(W'_{cls}\), respectively. The generalised intersection over union loss helps in enhancing the precision of the tracker's 3D bounding box regression for better alignment.
\paragraph{Architectural Details of the Smoother.}
Noise from LiDAR detections often stems from environmental factors such as occlusions, reflective surfaces, and sensor artifacts. Therefore, it is essential to design pre-processing techniques that filter out unreliable detections while maintaining valuable object information.
The smoother serves as a pre-processing step to fine-tune detections and make up for some missed predictions in-between frames.

Thus, all detections over a period \(k\) of a moving window of length are denoted by \(\mathcal{D}^{\tau-k,\ldots,\tau+k}\).
The transformer encoder processes these input detections \(\mathcal{D}^{\tau-k,\ldots,\tau+k}\) for the input sequence $\mathcal{P}^{\tau-k,\ldots,\tau+k}$, to represent a relationship of each detection to other detections in the sequence in the form of embeddings. We define an embedding function, $\mathcal{E}^{\tau-k,\ldots,\tau+k} := \phi\left( \mathcal{D}^{\tau-k,\ldots,\tau+k} \right)$
that maps the detections \( \mathcal{D}^{\tau-k,\ldots,\tau+k} \) to a \( d \)-dimensional embedding, and the embedding set for all detections with number \(N\), within the time window.

The model encodes historical and future detections into a unified embedding space, containing \(\mathcal{E}^{\tau-k,\ldots,\tau+k}\), using a multi-head self-attention transformer encoder mechanism allowing it to learn long-term dependencies and patterns as shown in Figure \ref{fig:overall} (b). The generated embeddings are used to create object queries \(Q^\tau_{1,\ldots,L}\) and its corresponding positional encoding \(P^\tau_{1,\ldots,L}\) using a Top-K selection filter mechanism used by~\cite{Pinto2022CanDL}, which selects the top scoring embeddings. We denote the number of tokens selected by Top-K as $L$.
Each output query embedding \( Q^\tau_{i} \in \mathbb{R}^{d} \), $i=1,\ldots,L$, is computed as a non-linear transformation: 
\begin{equation}Q^\tau_{i} = \sigma(W_2 \text{ReLU}(W_1 \mathcal{E}^t_i + b_1) + b_2), \; \label{eq:3} \end{equation}
where \( W_1 \) and \( W_2 \) are learnable transformation matrices, \( b_1 \) and \( b_2 \) are learnable bias terms, \( \sigma \) is a nonlinearity (ReLU). 
Finally, the set of query embeddings at frame \( \tau \) is, $Q^\tau_{1,\ldots,L} = \left( Q^\tau_1, \ldots , Q^\tau_L \right) \in \mathbb{R}^{L \times d}$.
The selected query embeddings is used by the DETR decoder with self-attention layers to fine-tune the bounding boxes, object classifications and association scores of the current frame improving the stability of detections. 
Finally, we apply a feed forward network to regress the objects $\mathcal{S}^\tau$ belonging to the current frame $\tau$.

\paragraph{Architectural Details of the Tracker.}

Our proposed tracking method integrates a trans\-former-based architecture module for both online and offline multiple object tracking in the LiDAR domain.
The core of our approach is a transformer architecture designed to accept tracked objects from the previous frame $\tau-1$, detected objects from the current frame $\tau$ and features of the point clouds, and output tracked objects in the current frame. 

For each frame, we start by taking the smoothed detections $\mathcal{S}^\tau$ and comparing (matching) them to the tracks from the previous frame $\mathcal{T}^{\tau-1}$. Any detections without a match are used to initialize new tracks, resulting in a set of track queries $\mathcal{Q'}^{\tau}$ that include both tracks from the previous frame and newly-proposed tracks from the detections. 
The tracker accepts as input the track queries  $\mathcal{Q'}^{\tau}$, the detected objects $\mathcal{S}^\tau$ and features $\mathcal{F}^{\tau-k+1, \ldots, \tau} = (\mathcal{F}^{\tau-k+1},\ldots,\mathcal{F}^{\tau})$ extracted from the point clouds to generate outputs $\mathcal{T}^{\tau}$ which are tracked objects at time $\tau$.
\begin{equation}
 \mathcal{T}^{\tau} = g(\mathcal{S}^\tau, \mathcal{Q'}^{\tau}, \mathcal{F}^{\tau-k+1, \ldots, \tau})
\end{equation}
At its core, the tracker is simply an attention block where the queries are the track queries $\mathcal{Q}^{\tau}$, the keys are the smoothed detections $S^\tau$ and the values are the point cloud features $\mathcal{F}^{\tau-k+1,\ldots,\tau}$. To achieve this, we first need to apply an encoder to the track queries $\mathcal{Q'}_{feat}^{\tau}=\mathit{encoder}(\mathcal{Q'}^{\tau})$ and the detected objects $\mathcal{S}_{feat}^{\tau}= \mathit{encoder}(\mathcal{S}^{\tau})$:
\begin{equation}
\mathcal{T}_{\mathit{feat}}^{\tau} = \text{Attention}(Q'^\tau_{\mathit{feat}}, \mathcal{S}_{\mathit{feat}}^\tau, \mathcal{F}^{\tau-k+1, \ldots, \tau}) 
\end{equation}
the attention block essentially searches for detection features ($\mathcal{S}_{\mathit{feat}}^{\tau}$) which are similar to our track queries (features), with the point cloud features provided as context, and outputs them as the new track features. From these features we decode the tracked objects in the current frame as $\mathcal{T}^{\tau} = \mathit{decoder}(\mathcal{T}_{\mathit{feat}}^{\tau})$.
The $\mathit{encoder}$ maps $\mathcal{S}^{\tau} \to \mathcal{S}_{\mathit{feat}}^{\tau}$ by means of a multi-layer perceptron (MLP) and positional encoding for spatial awareness. 
The $\mathit{decoder}$ also has an MLP which is applied to the output new track query: $\mathcal{T}^\tau = \sigma(W_b \mathcal{T}_{\mathit{feat}}^{\tau} + b_b)$, where $W_b$ and $b_b$ are learnable parameters and $\sigma$ is the sigmoid function, ensuring normalized outputs.

After decoding the features back into tracks $\mathcal{T}^{\tau}$, we filter out tracks with a low predicted confidence. The high-confidence tracks comprise the output of the tracker, and are also fed forward as track queries for the next frame.
The decoded object from the same query should be representing the same object across frames, thus forming a whole tracklet. To train the tracker, we need to assign one target ground-truth object for each query in each frame, and represent the assigned ground-truth object as the regression target for each object. Our mapping is defined as a one-to-one mapping where every predicted track is assumed to have only one corresponding ground-truth. In frames where the tracks exceed the ground-truth, we pad ground-truth with empty/null objects to ensure the one-to-one mapping is achieved. This streamlined design reduces computational overhead while improving the robustness and efficiency of LiDAR-based MOT.
Utilizing track queries, we are able to better handle data associations thereby tracking long-term with less identity mismatches~\cite{PointTransformer_2021_ICCV}.

To obtain the point cloud features, we apply the VoxelNet~\cite{voxelnet} voxelization to a stack of point clouds in the range $\tau-k+1, \ldots , \tau$. The weights of the feature encoder are also learned during training. These point cloud features provide context to the attention blocks so that an object which appears in multiple frames can be tracked across these frames based on the extracted LiDAR-based features.

\paragraph{Handling Occlusions and Ambiguities.}
We maintain a history buffer \cite{motr_camera} for each tracked object, storing information from past frames. This buffer enables predictive tracking when objects temporarily disappear due to occlusion or sensor limitations. When new detections emerge in subsequent frames, their state is fused with historical data to maintain continuity. The fusion process prioritizes recent observations while incorporating historical states to prevent drift and abrupt positional changes.
By aggregating information from multiple (10) LiDAR sweeps, the tracker can partially ``see'' occluded objects from alternate viewpoints. This ensures that tracking remains robust even in complex scenes.
The model integrates both appearance (3D shape) and motion (position, and trajectory) information over time to differentiate objects that may look similar but behave differently.

\begin{table*}[t!]
  \centering
  \scriptsize
  \setlength{\tabcolsep}{1.0pt}

  \begin{center}
  \resizebox{\textwidth}{!}{
    \begin{tabular}{lccccccccccccc}
      \toprule
      Method                                              &   & aMOTA$\uparrow$   & aMOTP$\downarrow$ &  & MOTA$\uparrow$ &MOTP $\uparrow$  & MT$\uparrow$  & ML$\downarrow$   & TP$\uparrow$   & FP$\downarrow$ & FN$\downarrow$    & IDS$\downarrow$ & FRAG$\downarrow$ \\
      \hline
    MotionTrack-L~\cite{Zhang2023MotionTrackET}  & & 0.51&0.99& &0.48 &0.30&3723&1567&& & &9705&\\

      CenterPoint~\cite{Centerpoint}                &  & 0.638             & 0.555             &  & 0.537 &0.284            & 5584          & 1681             & 95877          & 18612          & 22928             & 760             & 529              \\
       SimpleTrack~\cite{simpletrack}                &  & 0.668             & 0.550             &  & 0.537 &0.284            & 5584          & 1681             & 95877          & 18612          & 22928             & 575             & 529              \\
      UG3DMOT~  \cite{ug3dmot}          &  & 0.668             & 0.538            &  & 0.549&\underline{0.310}             & 5468          & 1776             & 95704          & 19401 & 22955             & 906             & 653              \\
    ImmortalTrack~\cite{Wang2021ImmortalTT} &&0.677 &0.599&&0.572& 0.285&
    5565&1669& 97584&18012&21661&\underline{320}&477\\
       
      3DMOTFormer~\cite{3dMotformer}     &  & 0.682             & 0.496             &  & 0.556&0.297             & 5466          & 1896    & 95790          & 18322          & 23357             & 438             & 529              \\

       NEBP~~\cite{NEBP2022}            &  & 0.683             & 0.624             &  & 0.584 &0.300& 5428          & 1993             & 97367          & 16773          & 21971             & \textbf{227}             & \textbf{299}  \\ 
      ShaSTA~~\cite{2023shasta}      &  & 0.696 & 0.540             &  & 0.578 & 0.295             & 5596          & 1813             & \textbf{97799}          & \textbf{16746}          & 21293 & 473             & \underline{356}              \\
       VoxelNeXt~ \cite{VoxelNext2023}    &  & 0.710             & 0.511             &  & 0.600 & 0.308          & 5529          & 1728 & 97075          & 18348          & 21836             & 654            & 537              \\
      FocalFormer3D  ~\cite{focalformer3d}          &  & \underline{0.715}             & 0.549             &  & \textbf{0.601}&0.309             & \underline{5615}          & \textbf{1550}             & 97535          & 16760          & \textbf{21142 }            & 888            & 810              \\
      
      \midrule
        Offline Track~\cite{offlineTrack}           &  & 0.671             & 0.522             &  & 0.553&0.296             & \textbf{5658}          & 1713             & 96617          & 16778          & 22378             & 570             & 592              \\
       \midrule
      
      LiDAR MOT-DETR (online)      & & 0.724   & 0.475   & & 0.594  & 0.310           &    5415      & 2063             &   97315        &  18446        &  21846           &    404       &    528         \\
      LiDAR MOT-DETR (offline)        &  & \textbf{0.726}    & \textbf{0.445}   &  & \underline{0.592}& \textbf{0.312}            & 5561          & 1892             & \underline{97593}          & 18695          & 21495             & 577             & 462              \\
      
      \hline
    \end{tabular}}
  \end{center}

  \caption{Overall results of LiDAR-only methods on the nuScenes test set based on Nuscenes Leaderboard. Best value highlighted in bold and second place underlined.}
  \label{tab:motmetrics}
\end{table*}

\section{Experiments}
LiDAR MOTR-DETR uses detections from off-the shelf object detectors and hence the performance of our method is influenced by the underlying detector. Thus, for a fair  comparison, we evaluate the performance with other tracking methods using the same object detector. We relied on two popular detectors used in previous papers~\cite{3dMotformer}, CenterPoint~\cite{Centerpoint} and Focalformer3D-L~\cite{focalformer3d}. We evaluate our results on both the nuScenes and Kitti public datasets. Further description of our dataset and data preprocessing steps can be found in the attached supplementary material. Although our smoother method primarily focused on an offline mode, the tracker is completely online as it uses point clouds from present and past frames only and the output of the smoother per key frame. To provide a fairer comparison with other online methods, we also trained an online variant of the smoother. All ablations were evaluated using the validation splits.

\subsection{Training and Inference.}
For better generalization and faster training, we pre-trained the smoother and tracker components on unannotated data, which was auto-ground-truthed using the CenterPoint model and a Kalman-based tracker.  This unannotated data is from an private dataset collected with a different lidar sensor (Velodyne VLS-128). We later used the pre-trained checkpoints as a starting point when training our model on nuScenes~\cite{caesar2020nuscenes} and KITTI~\cite{kitti2012CVPR}. For the CenterPoint and FocalFormer3D object detectors used in our results, we used the model checkpoints released by the authors~\cite{Centerpoint, focalformer3d}.\\ 
The smoother network is trained using detections with a temporal window of 15, ($k=7$) for 24 epochs by AdamW optimizer. The initial learning rate (LR) is set to 5e-4.  
This tracker is trained on multi-sweep LiDAR frames (sweep size = 10), plus the output detections of the smoother, for 32 epochs using the AdamW optimizer, LR of 4e-4 and bs of 6. 
The loss weights in equation \ref{eq:2}  are set as $W_{bbox}=0.5$, $W_{cls}=1.0$ with $\alpha = 0.5$ and $\gamma =2$. For the tracker, we set $W'_{bbox}=0.25$, $W'_{cls}=2.0$ with $\alpha = 0.25$ and $\gamma =2$. 
\\
\textbf{Inference.} 
During inference, our model processes raw inputs from multiple LiDAR sweeps, updates the track queries on a frame-by-frame basis, and outputs a complete set of object tracks.
The relevant tracks are selected using a confidence-based pruning method 
$\mathcal{T}^\tau = \{ t_i^\tau \mid C_i^\tau < \tau_{c}, t_i^\tau \in T^{\tau-1} \},$
where \( \tau_{c} \) is the termination confidence threshold (0.2) and $C$ is the track score. 
Inactive tracks are terminated when they exceed a predefined maximum of allowed frames. In this experiment, inactive tracks are terminated after 5 frames. This ensures that objects missing for several consecutive frames are eventually removed.

\begin{table*}[t!]
  \centering
  \scriptsize
  \setlength{\tabcolsep}{1.5pt}
  \begin{center}
    \begin{tabular}{lcccccccc}
      \toprule
      Method & Detector       &Modality         & mAP$\uparrow$   & mATE$\downarrow$ &  mASE$\downarrow$    & mAOE$\downarrow$  & mAAE$\downarrow$   & NDS$\uparrow$  \\
    \midrule
      CenterPoint &CenterPoint &L&0.564  & -                         &-  & -            & & 0.648      \\
      Ours (Online Smoother)~\ &CenterPoint&L& 0.637              &0.255  & \textbf{0.21.0}             & 0.364            &0.138  & 0.695                \\
      Ours (Offline Smoother) ~\ &CenterPoint& L&0.671& 0.253 & 0.214               &0.342  & 0.135             & 0.711  \\
      
        FocalFormer3D~\cite{focalformer3d}         &FocalFormer&L& 0.664    & -& -          &    -          &- & 0.709            \\          
      
      Ours (Online Smoother) &FocalFormer&L& 0.672   & \textbf{0.249}  & 0.245             & \textbf{0.331 }            & \textbf{0.125}  & \textbf{0.725}             \\
      Ours (Offline Smoother)~&FocalFormer&L & \textbf{0.683}   & 0.251  & 0.242             & 0.337             & 0.133  & 0.717 \\
       
      \bottomrule
      
    \end{tabular}
  \end{center}
  \caption{Smoother results on nuScenes validation set with temporal window of 15, when run on top of two different object detectors. L represents models trained using lidar pointclouds.}
  \label{tab:test_split}
\end{table*}

\begin{table}[!htb]
\resizebox{\textwidth}{!}{
\begin{minipage}{.5\linewidth}
  \centering
  \scriptsize
  \setlength{\tabcolsep}{1.5pt}
  \begin{center}
    \begin{tabular}{cccccc}
      \hline
    Total Frames & aMOTA$\uparrow$ & aMOTP$\downarrow$  & IDS$\downarrow$ & FRAG$\downarrow$ &mAP$\uparrow$\\
      \hline
      8   & 0.682       & 0.559  &  442      &       506      & 0.632    \\
      11  &   0.705     &   0.556        &     425      &     500       & 0.657         \\
      13  &   0.726        &  0.538        &   419     &      480        & 0.670             \\
      15 & \textbf{0.735}       & \textbf{0.523}   &     \textbf{407}      &        \textbf{474}     & \textbf{0.671}    \\
    \hline
    \end{tabular}
    \\(a) offline
  \end{center}
  \label{tab:sample_len_a}
  \end{minipage}

\begin{minipage}{.5\linewidth}
  \centering
  \scriptsize
  \setlength{\tabcolsep}{1.5pt}
  \begin{center}
    \begin{tabular}{cccccc}
      \hline
       Total Frames & aMOTA$\uparrow$ & aMOTP$\downarrow$  & IDS$\downarrow$ & FRAG$\downarrow$ &mAP$\uparrow$\\
      \hline
   
        8 &   0.677     &   0.557      &    457      &    502     &   0.629       \\
      11   & 0.698        & 0.542 &   428        &         498    & 0.635    \\ 
      13 &  0.713     &   0.532      &   424       &    497       &  0.634        \\
      15&  \textbf{0.729}     & \textbf{0.529} &  \textbf{410}       &   \textbf{462}        & \textbf{0.637}  \\
      
    \hline
    \end{tabular}\\
    (b) online
  \end{center}
  \label{tab:sample_len_b}
  \end{minipage}
}
\caption{Comparison of tracker performance with different temporal window sizes of Smoother. Results are generated using CenterPoint detections,  on nuScenes validation set in (a) offline and (b) online modes.}
\label{tab:sample_len}
\end{table}

\subsection{Primary Evaluation Metrics}
\begin{figure}
\begin{tabular}{cccc}
\bmvaHangBox{\fbox{\includegraphics[width=2.5cm]{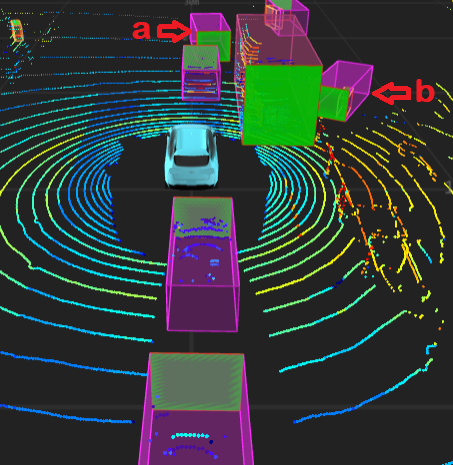}}}&
\bmvaHangBox{\fbox{\includegraphics[width=2.5cm]{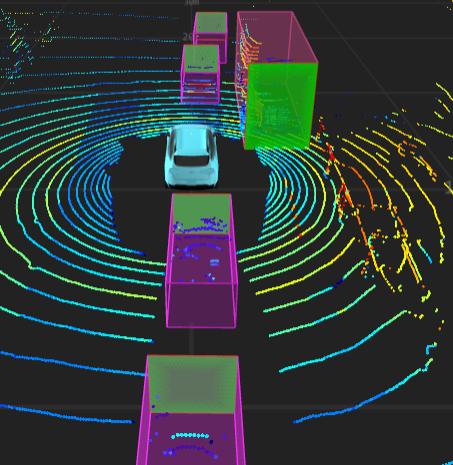}}}&
\bmvaHangBox{\fbox{\includegraphics[width=2.5cm]{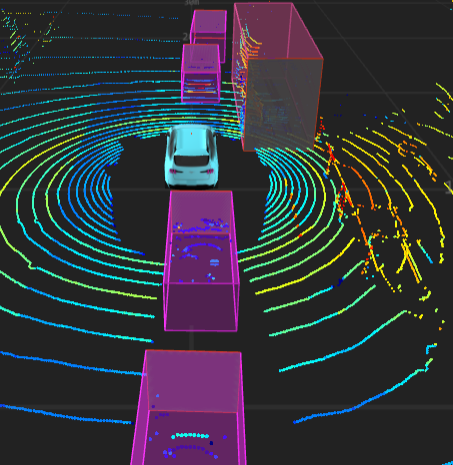}}}&
\bmvaHangBox{\fbox{\includegraphics[width=2.5cm]{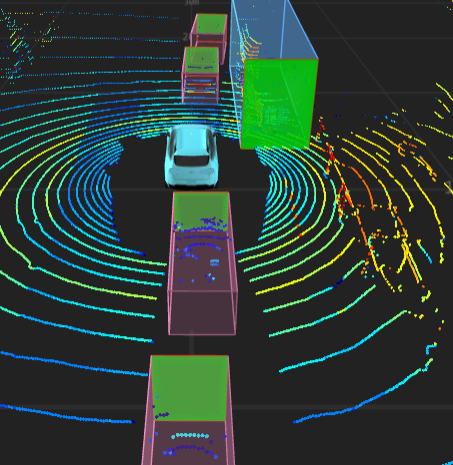}}}
\\
(detections)&(smoother)&(tracker)&(ground-truth)
\end{tabular}
\caption{Visual representation on nuScenes validation set. In (detections), we show arrow a, an object with wrong orientation and facing the wrong (opposite) direction from object detector. Arrow b shows false positive from detector. We recommend zooming in.}
\label{fig:demo}
\end{figure}

MOT are commonly evaluated on two main metrics: Average Multiple Object Track Accuracy (aMOTA), which measures how well the tracker detects objects in the scene, and Multiple Object Track Precision (aMOTP)~\cite{mota}, which measures how well the tracker's bounding boxes fit the objects. 
On Kitti, the performance is evaluated using Higher Order Tracking Accuracy (HOTA)~\cite{Hota2020IJCV}, both MOTA and MOTP~\cite{mota}. Additionally, we compare the different methods by identity switches (IDS/IDSW)~\cite{mota, mot20} and track fragmentation (FRAG)~\cite{mota}, as well as a breakdown of how many of the objects were tracked across at least 80\% of their lifetime (Mostly Tracked, MT) or at most 20\% of their lifetime (Mostly Lost, ML)~\cite{mot20}. Any tracks which lie in between the MT and ML are considered Partially Tracked (PT) objects.
Table \ref{tab:motmetrics} compares the performance of LiDAR MOT-DETR with baselines and state-of-the-art deep tracking methods on the nuScenes dataset. We observed that the aMOTP is significantly better than other methods, at 0.475 (compared to 0.549 for SOTA FocalFormer3D-L), while aMOTA remains slightly above SOTA. Thus, while LiDAR MOT-DETR's tracker accuracy is better than other methods (with an aMOTA of 0.724,  compared to 0.715 of SOTA), its main contribution is in its refinement of the predicted bounding boxes as seen in figures \ref{fig:demo1} and \ref{fig:demo}. Although we do not outperform LiDAR SOTA~\cite{castrack, Wang2024MCTrackAU, huang2025bitrackbidirectionaloffline3d, RobMOT, zhang2023lego, pctcnn} on KITTI dataset, we see a very competitive performance in HOTA of 0.822 in table \ref{tab:kitti_data}. \\
\\
\textbf{Smoother design choice -- offline vs.\ online.} 
We found that the offline (forward-peeking) variant of the smoother outperforms the online variant. It achieved an NDS score (nuScenes Detection Score -- weighted sum of multiple detection metrics \cite{motmetrics-nus}) of 0.711, as opposed to 0.695 for the online variant (table \ref{tab:test_split}). Both of these significantly outperform the 0.648 score of the CenterPoint detector itself.  Use cases such as auto ground-truthing, which do not need to run in real-time, would benefit from this increased performance.\\
\\
\textbf{Length of smoother temporal window.} We found that the performance of the tracker increases as we increase the size of the smoother's temporal window, in both the online and the offline (forward-peeking) variants, as observed in the aMOTA and aMOTP (table \ref{tab:sample_len}). When fixing the window size to 15 frames, the forward-peeking variant outperforms the online variant by 1.7 percent points (\,pp).
\begin{table}

\resizebox{0.96\textwidth}{!}{
\begin{minipage}{.44\linewidth}
  \scriptsize
  \setlength{\tabcolsep}{1pt}
    \begin{tabular}{lc|c|ccccc}
   
      \hline
        Name  &&Sensor   &HOTA &DetA &IDSW &MOTA \\
    \hline
        UG3DMOT~\cite{ug3dmot}& & L  &0.808  &0.782 & 13 & 0.866\\
        MCTRACK~\cite{Wang2024MCTrackAU}  &&L+C &0.839  & -& 3 &0.64\\
        BiTrack~\cite{huang2025bitrackbidirectionaloffline3d}  &&L+C &0.845  &\textbf{0.819} & 13 &0.878\\
        RobMOT~\cite{RobMOT} & & L  &0.863 &- &1 &0.915\\
        CasTrack~\cite{castrack} & & L  &0.932 &- &- &-\\
        PC-TCNN~\cite{pctcnn} & & L  &0.944 &- &3 &0.886\\
        LeGO~\cite{zhang2023lego}& & L &  \textbf{0.952} & -& 1 & 0.90\\
        \hline
        & &\\
         Ours (Online)&& L   & 0.852 &0.817 & 12 &0.913 \\
         Ours (Offline)&& L   & 0.894 &0.802& 8&0.916 &\\
        
    \hline
    \end{tabular}
    \caption{Results on KITTI validation dataset.\\
    }
  \label{tab:kitti_data}

\end{minipage}
\begin{minipage}{0.02\linewidth}
\tiny
    \begin{tabular}{c}
         \\
    \end{tabular}
\end{minipage}

\begin{minipage}{.5\linewidth}
    \scriptsize
    \setlength{\tabcolsep}{1pt}
    \begin{tabular}{l|c|cccc} 
      \hline
      Method                                              & Detector  & aMOTA$\uparrow$   & aMOTP$\downarrow$ &  IDS$\downarrow$ & FRAG$\downarrow$ \\
      \hline
     
      CenterPoint~\cite{Centerpoint}                & Centerpoint & 0.637             & 0.606            & -             & -              \\
      VoxelNeXt~~\cite{VoxelNext2023}     &End-to-End& 0.702            & 0.640             & 729             &            -   \\
      NEBP ~\cite{NEBP2022}  & CenterPoint&         0.708             &    -          &              172             & 271              \\
       3DMOTFormer~~\cite{NEBP2022}           & CenterPoint&0.712             & 0.515             & 341             & 436  \\ 
      ShaSTA~~\cite{2023shasta}      & CenterPoint & 0.728 &  -     &    -          &        -       \\
      Ours (w/o Smoother)      &CenterPoint  & 0.681 &    0.579      &  435            &501        \\
      Ours (Two-stage)           & CenterPoint & \underline{0.735}           &   0.523                & 407             & 474              \\
      \hline

      FocalFormer3D-L  ~\cite{focalformer3d}          & FocalFormer & 0.721             & - &     -        &        -       \\
      Ours (w/o Smoother)      & FocalFormer & 0.729 &    0.486         &  398            &485        \\
      Ours (Two-stage)           & FocalFormer& \textbf{0.752}           &   \textbf{0.479}           &   \textbf{332}    & \textbf{454}              \\
      \hline
      
    \end{tabular}

  \caption{Results on nuScenes validation set with and without smoother in comparison with other methods.}
  \label{tab:motmetrics-centerpoint}
\end{minipage}

}

\end{table}

\begin{table}
  \scriptsize
  \setlength{\tabcolsep}{2.0pt}

  \begin{center}
    \begin{tabular}{lc|c|c|cccc} 
      \hline
           & Detector &Pre-training &Smoother& aMOTA$\uparrow$   & aMOTP$\downarrow$ &  IDS$\downarrow$ & FRAG$\downarrow$ \\
      \hline
            &CenterPoint &\xmark & \xmark & 0.672 &   0.581       &    456         &   521     \\
            & CenterPoint & \xmark& \cmark  & 0.685           & 0.573                  &   438           &  493             \\
      
           &CenterPoint &\cmark & \xmark & 0.681 &    0.579      &  435            &501        \\
                & CenterPoint & \cmark& \cmark  & \underline{0.735}           &   0.523                & 407             & 474              \\
     
      \hline
               & FocalFormer-L &\xmark & \xmark   &  0.728            & 0.541 &     419       &        489       \\
                & FocalFormer-L &\xmark & \cmark   &     0.736         & 0.532 &    392       &        \underline{468 }      \\
            & FocalFormer-L & \cmark & \xmark & 0.729 &    0.486         &  398            &485        \\
               & FocalFormer-L & \cmark& \cmark & \textbf{0.752}           &   \textbf{0.479}           &   \textbf{332}    & \textbf{454}              \\
                
      \hline
      
    \end{tabular}
  \end{center}

  \caption{Ablation study on the nuScenes validation set; (1) with and without pre-training; (2) with and without smoother.}
  \label{tab:pretraining}
\end{table}

\paragraph{Effect of base object detector used.}
We evaluated our method on both FocalFormer3D \cite{focalformer3d}, a SOTA object detector, and CenterPoint \cite{Centerpoint}, an older object detector. We found that a large 8.4\,pp aMOTA difference between the detectors themselves was reduced to only 1.7\,pp post-tracking in (table \ref{tab:motmetrics-centerpoint}) with our two stage method. Therefore, the gains provided by LiDAR MOT-DETR are higher when used with older/underperforming object detectors.\\
\\
\textbf{Effect of smoother model.}
Additionally, we evaluated the performance of the tracker with and without the underlying smoother model. Table \ref{tab:motmetrics-centerpoint} shows that the performance gain provided when applying the smoother to the CenterPoint detections (5.4\,pp) is much larger than the gain when applying it to the FocalFormer3D detections (2.3\,pp). Thus, much of the gain which LiDAR MOT-DETR gives to the older object detectors comes from the smoothing step, not necessarily the tracker. This highlights the importance of the smoother model in our approach, as it helps maintain internal consistency of bounding boxes. An example of this is illustrated in Figure \ref{fig:demo}, where we see how poorly-detected bounding boxes from the raw detector are corrected by the smoother and tracker.\\
\\
\textbf{Effect of pre-training.}
In table \ref{tab:pretraining}, we demonstrate the advantage and performance boost we get as a result of pre-training our model (excluding the detector) on a large-scale auto ground-truthed proprietary dataset. When training our model solely on nuScenes, we see a 1.3\,pp improvement in aMOTA when incorporating the smoother instead of when not. However, we achieve an additional 5\,pp in aMOTA when pre-training our model on the proprietary dataset. This shows, although the proprietary dataset rather consists of automatically generated ground-truth, our smoother and tracker clearly benefit from large volumes of data.\\

In summary, we have observed that our two-stage smoother and tracker exceeds state-of-the-art tracker performance on the nuScenes dataset. In particular, it significantly outperforms SOTA methods in the aMOTP metric, which reflects the accuracy of the bounding box. We believe that the smoother, which integrates detections across a window temporal window, is a major contributor for this. The same method can be used  either in an online or forward-peeking variation; this allows offline use cases, such as auto ground-truthing, to take advantage of the increased performance in forward-peeking mode.

\section{Conclusion}
Neural networks have revolutionized the field of multi-object tracking, providing powerful tools for robust and accurate tracking in complex scenarios. 
We present a novel two-staged approach to multiple object tracking in the LiDAR domain, utilizing Deformable DETR-based architectures, which uses existing off-the-shelf object detectors. Our tracker outperforms current state-of-the-art LiDAR-based MOTs by 0.9\,pp aMOTA and 10.5\,pp aMOTP on nuScenes validation sets. without extensive manual parameter tuning or post-processing of the obtained tracks.
This shows that the integration of transformers allows for the effective modelling of temporal dependencies. Our extensive evaluation demonstrates that this approach outperforms state-of-the-art methods in both accuracy and precision.\\

\bibliography{egbib}

\begin{thebibliography}{59}
\providecommand{\natexlab}[1]{#1}
\providecommand{\url}[1]{\texttt{#1}}
\expandafter\ifx\csname urlstyle\endcsname\relax
  \providecommand{\doi}[1]{doi: #1}\else
  \providecommand{\doi}{doi: \begingroup \urlstyle{rm}\Url}\fi

\bibitem[Ahn and Cho(2022)]{videosurvillance}
Hyochang Ahn and Han-Jin Cho.
\newblock Research of multi-object detection and tracking using machine learning based on knowledge for video surveillance system.
\newblock \emph{Personal and Ubiquitous Computing}, 26\penalty0 (2):\penalty0 385--394, 2022.

\bibitem[Bahraini et~al.(2019)Bahraini, Rad, and Bozorg]{bahraini2019slam}
Masoud~S Bahraini, Ahmad~B Rad, and Mohammad Bozorg.
\newblock Slam in dynamic environments: A deep learning approach for moving object tracking using ml-ransac algorithm.
\newblock \emph{Sensors}, 19\penalty0 (17):\penalty0 3699, 2019.

\bibitem[Bergmann et~al.(2019)Bergmann, Meinhardt, and Leal-Taix{\'e}]{TWB_2d}
Philipp Bergmann, Tim Meinhardt, and Laura Leal-Taix{\'e}.
\newblock Tracking without bells and whistles.
\newblock In \emph{Proceedings of the IEEE/CVF International Conference on Computer Vision}, pages 941--951, 2019.

\bibitem[Bernardin and Stiefelhagen(2008)]{mota}
Keni Bernardin and Rainer Stiefelhagen.
\newblock Evaluating multiple object tracking performance: the clear mot metrics.
\newblock \emph{EURASIP Journal on Image and Video Processing}, 2008:\penalty0 1--10, 2008.

\bibitem[Caesar et~al.(2020)Caesar, Bankiti, Lang, Vora, Liong, Xu, Krishnan, Pan, Baldan, and Beijbom]{caesar2020nuscenes}
Holger Caesar, Varun Bankiti, Alex~H Lang, Sourabh Vora, Venice~Erin Liong, Qiang Xu, Anush Krishnan, Yu~Pan, Giancarlo Baldan, and Oscar Beijbom.
\newblock nuscenes: A multimodal dataset for autonomous driving.
\newblock In \emph{Proceedings of the IEEE/CVF conference on computer vision and pattern recognition}, pages 11621--11631, 2020.

\bibitem[Carion et~al.(2020)Carion, Massa, Synnaeve, Usunier, Kirillov, and Zagoruyko]{detrE2E}
Nicolas Carion, Francisco Massa, Gabriel Synnaeve, Nicolas Usunier, Alexander Kirillov, and Sergey Zagoruyko.
\newblock End-to-end object detection with transformers.
\newblock \emph{CoRR}, abs/2005.12872, 2020.

\bibitem[Chen et~al.(2023{\natexlab{a}})Chen, Shi, Zhang, Zhu, Wang, Cheung, See, and Li]{chen2023trajectoryformer3dobjecttracking}
Xuesong Chen, Shaoshuai Shi, Chao Zhang, Benjin Zhu, Qiang Wang, Ka~Chun Cheung, Simon See, and Hongsheng Li.
\newblock Trajectoryformer: 3d object tracking transformer with predictive trajectory hypotheses, 2023{\natexlab{a}}.

\bibitem[Chen et~al.(2023{\natexlab{b}})Chen, Yu, Chen, Lan, Anandkumar, Jia, and Alvarez]{focalformer3d}
Yilun Chen, Zhiding Yu, Yukang Chen, Shiyi Lan, Anima Anandkumar, Jiaya Jia, and Jose~M Alvarez.
\newblock Focalformer3d: focusing on hard instance for 3d object detection.
\newblock In \emph{Proceedings of the IEEE/CVF International Conference on Computer Vision}, pages 8394--8405, 2023{\natexlab{b}}.

\bibitem[Chen et~al.(2023{\natexlab{c}})Chen, Liu, Zhang, Qi, and Jia]{VoxelNext2023}
Yukang Chen, Jianhui Liu, Xiangyu Zhang, Xiaojuan Qi, and Jiaya Jia.
\newblock Voxelnext: Fully sparse voxelnet for 3d object detection and tracking.
\newblock In \emph{Proceedings of the IEEE/CVF Conference on Computer Vision and Pattern Recognition (CVPR)}, pages 21674--21683, June 2023{\natexlab{c}}.

\bibitem[Dendorfer(2020)]{mot20}
P~Dendorfer.
\newblock Mot20: A benchmark for multi object tracking in crowded scenes.
\newblock \emph{arXiv preprint arXiv:2003.09003}, 2020.

\bibitem[Ding et~al.(2023)Ding, Rehder, Schneider, Cordts, and Gall]{3dMotformer}
Shuxiao Ding, Eike Rehder, Lukas Schneider, Marius Cordts, and Juergen Gall.
\newblock 3dmotformer: Graph transformer for online 3d multi-object tracking.
\newblock In \emph{Proceedings of the IEEE/CVF International Conference on Computer Vision (ICCV)}, pages 9784--9794, October 2023.

\bibitem[Du et~al.(2023)Du, Zhao, Song, Zhao, Su, Gong, and Meng]{du2023strongsort}
Yunhao Du, Zhicheng Zhao, Yang Song, Yanyun Zhao, Fei Su, Tao Gong, and Hongying Meng.
\newblock Strongsort: Make deepsort great again.
\newblock \emph{IEEE Transactions on Multimedia}, 25:\penalty0 8725--8737, 2023.

\bibitem[Erabati and Araujo(2023)]{li3detr}
Gopi~Krishna Erabati and Helder Araujo.
\newblock Li3detr: A lidar based 3d detection transformer.
\newblock In \emph{Proceedings of the IEEE/CVF Winter Conference on Applications of Computer Vision}, pages 4250--4259, 2023.

\bibitem[Geiger et~al.(2012)Geiger, Lenz, and Urtasun]{kitti2012CVPR}
Andreas Geiger, Philip Lenz, and Raquel Urtasun.
\newblock Are we ready for autonomous driving? the kitti vision benchmark suite.
\newblock In \emph{Conference on Computer Vision and Pattern Recognition (CVPR)}, 2012.

\bibitem[Hasan et~al.(2022{\natexlab{a}})Hasan, Hanawa, Goto, Suzuki, Fukuda, Kuno, and Kobayashi]{contreview2022}
Mahmudul Hasan, Junichi Hanawa, Riku Goto, Ryota Suzuki, Hisato Fukuda, Yoshinori Kuno, and Yoshinori Kobayashi.
\newblock Lidar-based detection, tracking, and property estimation: A contemporary review.
\newblock \emph{Neurocomputing}, 506:\penalty0 393--405, 2022{\natexlab{a}}.

\bibitem[Hasan et~al.(2022{\natexlab{b}})Hasan, Hanawa, Goto, Suzuki, Fukuda, Kuno, and Kobayashi]{lidarreview}
Mahmudul Hasan, Junichi Hanawa, Riku Goto, Ryota Suzuki, Hisato Fukuda, Yoshinori Kuno, and Yoshinori Kobayashi.
\newblock Lidar-based detection, tracking, and property estimation: A contemporary review.
\newblock \emph{Neurocomputing}, 506:\penalty0 393--405, 2022{\natexlab{b}}.
\newblock ISSN 0925-2312.
\newblock \doi{https://doi.org/10.1016/j.neucom.2022.07.087}.

\bibitem[He et~al.(2024)He, Fu, Wang, and Wang]{ug3dmot}
Jiawei He, Chunyun Fu, Xiyang Wang, and Jianwen Wang.
\newblock 3d multi-object tracking based on informatic divergence-guided data association.
\newblock \emph{Signal Processing}, 222:\penalty0 109544, 2024.

\bibitem[Hu et~al.(2021)Hu, Cai, Mlodzikowski, and Huang]{motionchallenge}
Peixuan Hu, Ruichen Cai, Witold Mlodzikowski, and Guan Huang.
\newblock Monocular 3d object detection and tracking using class-aware motion estimation.
\newblock In \emph{Proceedings of the IEEE/CVF International Conference on Computer Vision}, pages 864--874, 2021.

\bibitem[Huang et~al.(2025)Huang, Chen, Zhang, and Hao]{huang2025bitrackbidirectionaloffline3d}
Kemiao Huang, Yinqi Chen, Meiying Zhang, and Qi~Hao.
\newblock Bitrack: Bidirectional offline 3d multi-object tracking using camera-lidar data, 2025.
\newblock URL \url{https://arxiv.org/abs/2406.18414}.

\bibitem[Lee et~al.(2022)Lee, Lee, Shin, and Yi]{lee2022moving}
Hojoon Lee, Hyunsung Lee, Donghoon Shin, and Kyongsu Yi.
\newblock Moving objects tracking based on geometric model-free approach with particle filter using automotive lidar.
\newblock \emph{IEEE Transactions on Intelligent Transportation Systems}, 23\penalty0 (10):\penalty0 17863--17872, 2022.

\bibitem[Li et~al.(2024)Li, Liu, Zhao, Wu, Wu, and Gao]{FastPoly}
Xiaoyu Li, Dedong Liu, Lijun Zhao, Yitao Wu, Xian Wu, and Jinghan Gao.
\newblock Fast-poly: A fast polyhedral framework for 3d multi-object tracking.
\newblock \emph{arXiv preprint arXiv:2403.13443}, 2024.

\bibitem[Liang and Meyer(2022)]{NEBP2022}
Mingchao Liang and Florian Meyer.
\newblock Neural enhanced belief propagation for data association in multiobject tracking.
\newblock In \emph{2022 25th International Conference on Information Fusion (FUSION)}, pages 1--7. IEEE, 2022.

\bibitem[Lin et~al.(2020)Lin, Goyal, Girshick, He, and Dollár]{focalloss}
Tsung-Yi Lin, Priya Goyal, Ross Girshick, Kaiming He, and Piotr Dollár.
\newblock Focal loss for dense object detection.
\newblock \emph{IEEE Transactions on Pattern Analysis and Machine Intelligence}, 42\penalty0 (2):\penalty0 318--327, 2020.
\newblock \doi{10.1109/TPAMI.2018.2858826}.

\bibitem[Liu and Caesar(2024)]{offlineTrack}
Xianzhong Liu and Holger Caesar.
\newblock Offline tracking with object permanence.
\newblock In \emph{2024 IEEE Intelligent Vehicles Symposium (IV)}, pages 1272--1279. IEEE, 2024.

\bibitem[Luiten et~al.(2020)Luiten, Osep, Dendorfer, Torr, Geiger, Leal-Taixe, and Leibe]{Hota2020IJCV}
Jonathon Luiten, Aljosa Osep, Patrick Dendorfer, Philip Torr, Andreas Geiger, Laura Leal-Taixe, and Bastian Leibe.
\newblock Hota: A higher order metric for evaluating multi-object tracking.
\newblock \emph{International Journal of Computer Vision (IJCV)}, 2020.

\bibitem[Nagy et~al.(2024)Nagy, Werghi, Hassan, Dias, and Khonji]{RobMOT}
Mohamed Nagy, Naoufel Werghi, Bilal Hassan, Jorge Dias, and Majid Khonji.
\newblock Robmot: Robust 3d multi-object tracking by observational noise and state estimation drift mitigation on lidar pointcloud.
\newblock \emph{arXiv preprint arXiv:2405.11536}, 2024.

\bibitem[Pan et~al.(2024)Pan, Ding, Zhong, and Lu]{pan2024ratrack}
Zhijun Pan, Fangqiang Ding, Hantao Zhong, and Chris~Xiaoxuan Lu.
\newblock Ratrack: Moving object detection and tracking with 4d radar point cloud.
\newblock In \emph{2024 IEEE International Conference on Robotics and Automation (ICRA)}, pages 4480--4487. IEEE, 2024.

\bibitem[Pang et~al.(2022)Pang, Li, and Wang]{simpletrack}
Ziqi Pang, Zhichao Li, and Naiyan Wang.
\newblock Simpletrack: Understanding and rethinking 3d multi-object tracking.
\newblock In \emph{European Conference on Computer Vision}, pages 680--696. Springer, 2022.

\bibitem[Pinto et~al.(2022)Pinto, Hess, Ljungbergh, Xia, Wymeersch, and Svensson]{Pinto2022CanDL}
Juliano Pinto, Georg Hess, William Ljungbergh, Yuxuan Xia, Henk Wymeersch, and Lennart Svensson.
\newblock Can deep learning be applied to model-based multi-object tracking?
\newblock \emph{ArXiv}, abs/2202.07909, 2022.

\bibitem[Qi et~al.(2021)Qi, Zhou, Najibi, Sun, Vo, Deng, and Anguelov]{Qi2021Offboard3O}
C.~Qi, Yin Zhou, Mahyar Najibi, Pei Sun, Khoa~T. Vo, Boyang Deng, and Dragomir Anguelov.
\newblock Offboard 3d object detection from point cloud sequences.
\newblock \emph{2021 IEEE/CVF Conference on Computer Vision and Pattern Recognition (CVPR)}, pages 6130--6140, 2021.

\bibitem[Rezatofighi et~al.(2019)Rezatofighi, Tsoi, Gwak, Sadeghian, Reid, and Savarese]{giou}
Hamid Rezatofighi, Nathan Tsoi, JunYoung Gwak, Amir Sadeghian, Ian Reid, and Silvio Savarese.
\newblock Generalized intersection over union: A metric and a loss for bounding box regression, 2019.
\newblock URL \url{https://arxiv.org/abs/1902.09630}.

\bibitem[Ruppel et~al.(2022)Ruppel, Faion, Gläser, and Dietmayer]{pctrack}
Felicia Ruppel, Florian Faion, Claudius Gläser, and Klaus Dietmayer.
\newblock Transformers for multi-object tracking on point clouds.
\newblock In \emph{2022 IEEE Intelligent Vehicles Symposium (IV)}, pages 852--859, 2022.
\newblock \doi{10.1109/IV51971.2022.9827344}.

\bibitem[Sadjadpour et~al.(2023)Sadjadpour, Li, Ambrus, and Bohg]{2023shasta}
Tara Sadjadpour, Jie Li, Rares Ambrus, and Jeannette Bohg.
\newblock Shasta: Modeling shape and spatio-temporal affinities for 3d multi-object tracking.
\newblock \emph{IEEE Robotics and Automation Letters}, 2023.

\bibitem[Sun et~al.(2020)Sun, Cao, Jiang, Zhang, Luo, Wang, and Liu]{sun2020transtrack}
Peize Sun, Yibing Cao, Yi~Jiang, Rufeng Zhang, Ping Luo, Xiaogang Wang, and Wenyu Liu.
\newblock Transtrack: Multiple-object tracking with transformer.
\newblock \emph{arXiv preprint arXiv:2012.15460}, 2020.

\bibitem[Tian et~al.(2020)Tian, Lauer, and Chen]{trafficTrack}
Wei Tian, Martin Lauer, and Long Chen.
\newblock Online multi-object tracking using joint domain information in traffic scenarios.
\newblock \emph{IEEE Transactions on Intelligent Transportation Systems}, 21\penalty0 (1):\penalty0 374--384, 2020.
\newblock \doi{10.1109/TITS.2019.2892413}.

\bibitem[Vaswani(2017)]{vaswani2017attention}
A~Vaswani.
\newblock Attention is all you need.
\newblock \emph{Advances in Neural Information Processing Systems}, 2017.

\bibitem[Wang et~al.(2020)Wang, Sun, Hoang, Harakeh, and Waslander]{wang2020pointtracknet}
Hao Wang, Peiyun Sun, Thien Hoang, Ali Harakeh, and Steven~L. Waslander.
\newblock Pointtracknet: An end-to-end network for 3d object detection and tracking from point clouds.
\newblock \emph{arXiv preprint arXiv:2012.02353}, 2020.

\bibitem[Wang et~al.(2021)Wang, Chen, Pang, Wang, and Zhang]{Wang2021ImmortalTT}
Qitai Wang, Yuntao Chen, Ziqi Pang, Naiyan Wang, and Zhaoxiang Zhang.
\newblock Immortal tracker: Tracklet never dies.
\newblock \emph{ArXiv}, abs/2111.13672, 2021.
\newblock URL \url{https://api.semanticscholar.org/CorpusID:244709700}.

\bibitem[Wang et~al.(2023{\natexlab{a}})Wang, Liu, Wang, Li, and Zhang]{wang2023exploringobjectcentrictemporalmodeling}
Shihao Wang, Yingfei Liu, Tiancai Wang, Ying Li, and Xiangyu Zhang.
\newblock Exploring object-centric temporal modeling for efficient multi-view 3d object detection, 2023{\natexlab{a}}.

\bibitem[Wang et~al.(2023{\natexlab{b}})Wang, Fu, He, Huang, Meng, Zhang, Zhou, Xu, and Zhang]{twodetector}
Xiyang Wang, Chunyun Fu, Jiawei He, Mingguang Huang, Ting Meng, Siyu Zhang, Hangning Zhou, Ziyao Xu, and Chi Zhang.
\newblock You only need two detectors to achieve multi-modal 3d multi-object tracking.
\newblock \emph{arXiv preprint arXiv:2304.08709}, 2023{\natexlab{b}}.

\bibitem[Wang et~al.(2024)Wang, Qi, Zhao, Zhou, Zhang, Wang, Tu, Guo, Zhao, Li, and Yang]{Wang2024MCTrackAU}
Xiyang Wang, Shouzheng Qi, Jieyou Zhao, Hangning Zhou, Siyu Zhang, Guoan Wang, Kai Tu, Songlin Guo, Jianbo Zhao, Jian Li, and Mu~Yang.
\newblock Mctrack: A unified 3d multi-object tracking framework for autonomous driving.
\newblock \emph{ArXiv}, abs/2409.16149, 2024.

\bibitem[Wang et~al.(2019)Wang, Chao, Garg, Hariharan, Campbell, and Weinberger]{depthchallenge}
Yan Wang, Wei-Lun Chao, Divyansh Garg, Bharath Hariharan, Mark Campbell, and Kilian~Q Weinberger.
\newblock Pseudo-lidar from visual depth estimation: Bridging the gap in 3d object detection for autonomous driving.
\newblock In \emph{Proceedings of the IEEE/CVF Conference on Computer Vision and Pattern Recognition}, pages 8445--8453, 2019.

\bibitem[Weng et~al.(2020{\natexlab{a}})Weng, Wang, Held, and Kitani]{motmetrics-nus}
Xinshuo Weng, Jianren Wang, David Held, and Kris Kitani.
\newblock 3d multi-object tracking: A baseline and new evaluation metrics.
\newblock In \emph{2020 IEEE/RSJ International Conference on Intelligent Robots and Systems (IROS)}, pages 10359--10366, 2020{\natexlab{a}}.
\newblock \doi{10.1109/IROS45743.2020.9341164}.

\bibitem[Weng et~al.(2020{\natexlab{b}})Weng, Wang, Man, and Kitani]{weng2020gnn3dmot}
Xinshuo Weng, Yongxin Wang, Yunze Man, and Kris~M Kitani.
\newblock Gnn3dmot: Graph neural network for 3d multi-object tracking with 2d-3d multi-feature learning.
\newblock In \emph{Proceedings of the IEEE/CVF Conference on Computer Vision and Pattern Recognition}, pages 6499--6508, 2020{\natexlab{b}}.

\bibitem[Wojke et~al.(2017)Wojke, Bewley, and Paulus]{deepsort}
Nicolai Wojke, Alex Bewley, and Dietrich Paulus.
\newblock Simple online and realtime tracking with a deep association metric.
\newblock \emph{ICIP}, pages 3645--3649, 2017.

\bibitem[Wu et~al.(2021)Wu, Li, Wen, Li, Fan, and Wang]{pctcnn}
Hai Wu, Qing Li, Chenglu Wen, Xin Li, Xiaoliang Fan, and Cheng Wang.
\newblock Tracklet proposal network for multi-object tracking on point clouds.
\newblock In \emph{IJCAI}, pages 1165--1171, 2021.

\bibitem[Wu et~al.(2022)Wu, Deng, Wen, Li, Wang, and Li]{castrack}
Hai Wu, Jinhao Deng, Chenglu Wen, Xin Li, Cheng Wang, and Jonathan Li.
\newblock Casa: A cascade attention network for 3-d object detection from lidar point clouds.
\newblock \emph{IEEE Transactions on Geoscience and Remote Sensing}, 60:\penalty0 1--11, 2022.
\newblock \doi{10.1109/TGRS.2022.3203163}.

\bibitem[Yin et~al.(2020)Yin, Zhou, and Krahenbuhl]{yin2020lidar}
Tianwei Yin, Xingyi Zhou, and Philipp Krahenbuhl.
\newblock Lidar-based online 3d video object detection with graph-based message passing and spatiotemporal transformer attention.
\newblock In \emph{Proceedings of the IEEE/CVF Conference on Computer Vision and Pattern Recognition}, pages 11495--11504, 2020.

\bibitem[Yin et~al.(2021)Yin, Zhou, and Krahenbuhl]{Centerpoint}
Tianwei Yin, Xingyi Zhou, and Philipp Krahenbuhl.
\newblock Center-based 3d object detection and tracking.
\newblock In \emph{Proceedings of the IEEE/CVF Conference on Computer Vision and Pattern Recognition (CVPR)}, pages 11784--11793, June 2021.

\bibitem[Zeller et~al.(2024)Zeller, Herraez, Behley, Heidingsfeld, and Stachniss]{zeller2024radar}
Matthias Zeller, Daniel~Casado Herraez, Jens Behley, Michael Heidingsfeld, and Cyrill Stachniss.
\newblock Radar tracker: Moving instance tracking in sparse and noisy radar point clouds.
\newblock In \emph{2024 IEEE International Conference on Robotics and Automation (ICRA)}, pages 16170--16177. IEEE, 2024.

\bibitem[Zeng et~al.(2022)Zeng, Dong, Zhang, Wang, Zhang, and Wei]{motr_camera}
Fangao Zeng, Bin Dong, Yuang Zhang, Tiancai Wang, Xiangyu Zhang, and Yichen Wei.
\newblock Motr: End-to-end multiple-object tracking with transformer.
\newblock In \emph{European Conference on Computer Vision}, pages 659--675. Springer, 2022.

\bibitem[Zhang et~al.(2023{\natexlab{a}})Zhang, Zhang, Guo, Chen, and Happold]{Zhang2023MotionTrackET}
Ce~Zhang, Chengjie Zhang, Yiluan Guo, Lingji Chen, and Michael Happold.
\newblock Motiontrack: End-to-end transformer-based multi-object tracking with lidar-camera fusion.
\newblock \emph{2023 IEEE/CVF Conference on Computer Vision and Pattern Recognition Workshops (CVPRW)}, pages 151--160, 2023{\natexlab{a}}.
\newblock URL \url{https://api.semanticscholar.org/CorpusID:259286967}.

\bibitem[Zhang et~al.(2023{\natexlab{b}})Zhang, Gao, Xiao, and Fan]{zhang2023animaltrack}
Libo Zhang, Junyuan Gao, Zhen Xiao, and Heng Fan.
\newblock Animaltrack: A benchmark for multi-animal tracking in the wild.
\newblock \emph{International Journal of Computer Vision}, 131\penalty0 (2):\penalty0 496--513, 2023{\natexlab{b}}.

\bibitem[Zhang et~al.(2022)Zhang, Chen, Wang, Wang, and Zhao]{Mutr3d_camera}
Tianyuan Zhang, Xuanyao Chen, Yue Wang, Yilun Wang, and Hang Zhao.
\newblock Mutr3d: A multi-camera tracking framework via 3d-to-2d queries.
\newblock In \emph{Proceedings of the IEEE/CVF Conference on Computer Vision and Pattern Recognition (CVPR) Workshops}, pages 4537--4546, June 2022.

\bibitem[Zhang et~al.(2021)Zhang, Wang, Wang, and Liu]{zhang2020fairmot}
Yifu Zhang, Chunyu Wang, Xinggang Wang, and Wenjun Liu.
\newblock Fairmot: On the fairness of detection and re-identification in multiple object tracking.
\newblock \emph{International Journal of Computer Vision}, 129\penalty0 (11):\penalty0 3069--3087, 2021.

\bibitem[Zhang et~al.(2023{\natexlab{c}})Zhang, Liu, Xia, Huang, Han, and Liu]{zhang2023lego}
Zhenrong Zhang, Jianan Liu, Yuxuan Xia, Tao Huang, Qing-Long Han, and Hongbin Liu.
\newblock Lego: Learning and graph-optimized modular tracker for online multi-object tracking with point clouds.
\newblock \emph{arXiv preprint arXiv:2308.09908}, 2023{\natexlab{c}}.

\bibitem[Zhao et~al.(2021)Zhao, Jiang, Jia, Torr, and Koltun]{PointTransformer_2021_ICCV}
Hengshuang Zhao, Li~Jiang, Jiaya Jia, Philip~H.S. Torr, and Vladlen Koltun.
\newblock Point transformer.
\newblock In \emph{Proceedings of the IEEE/CVF International Conference on Computer Vision (ICCV)}, pages 16259--16268, October 2021.

\bibitem[Zhou et~al.(2020)Zhou, Koltun, and Kr{\"a}henb{\"u}hl]{TasPoints2D}
Xingyi Zhou, Vladlen Koltun, and Philipp Kr{\"a}henb{\"u}hl.
\newblock Tracking objects as points.
\newblock In \emph{European Conference on Computer Vision}, pages 474--490. Springer, 2020.

\bibitem[Zhou and Tuzel(2017)]{voxelnet}
Yin Zhou and Oncel Tuzel.
\newblock Voxelnet: End-to-end learning for point cloud based 3d object detection.
\newblock \emph{CoRR}, abs/1711.06396, 2017.

\end{thebibliography}


\title{ Supplementary for LiDAR MOT-DETR: A LiDAR-based Two-Stage
Transformer for 3D Multiple Object Tracking}


\addauthor{Martha Teiko Teye}{m.teye-hk@uni-wuppertal.de}{1,2}
\addauthor{Ori Maoz}{ori.maoz@aptiv.com}{2}
\addauthor{Matthias Rottmann}{matthias.rottmann@uos.de}{3}

\addinstitution{
School of Mathematics and Natural Sciences\\
University of Wuppertal, Germany\\
}
\addinstitution{
 Aptiv Services Deutschland GmbH\\
 Wuppertal, Germany\\
}
\addinstitution{
Department of Mathematics, Computer Science and Physics\\
 Osnabr\"{u}ck University, Germany\\
}

\runninghead{Teye, Maoz, Rottmann}{LiDAR MOT-DETR}

\def\eg{\emph{e.g}\bmvaOneDot}
\def\Eg{\emph{E.g}\bmvaOneDot}
\def\etal{\emph{et al}\bmvaOneDot}


\maketitle
\appendix
\section*{Supplementary Material}
\label{sec:supplementary}
The supplementary material for LiDAR MOT-DETR is organised as:
\begin{itemize}
    \item Section \ref{sec:A} gives additional details about the dataset used for training of our models.
    \item Section \ref{sec:B} shows experimental results of our models on Kitti Dataset.
    \item Section \ref{sec:C} discusses the runtime and memory usage.
    \item Section \ref{sec:D} demonstrates additional qualitative results of our two stage approach 
\end{itemize}

\section{Datasets and Training}
\label{sec:A}
We train our model on the nuScenes dataset ~\cite{caesar2020nuscenes} and KITTI dataset~\cite{kitti2012CVPR}. The nuScenes dataset comprises 850 training sequences of which 700 were used for training and 150 as validation and test set according to the nuScenes API split.Keyframes for the Lidar sensor are sampled at 2 FPS. We also use an unannotated private dataset mainly for model pre-training for the tracker. Labels for this dataset were autogenerated at a rate of 10 FPS This dataset uses different LiDAR sensors including HESAI Pandar 40 and Velodyne VLS-128 and consists of 232,281 lidar frames. The KITTI dataset is a comparatively small dataset which uses Velodyne HDL-64E lidar sensor and consists of 21 training sequences and 29 test sequences.
We rely on two main off-the-shelf object detectors, CenterPoint~\cite{Centerpoint} and FocalFormer3D~\cite{focalformer3d} to generate initial detections for our model. We also performed ablations using these two datasets to evaluate our model performance. 
The whole method is implemented in Python using PyTorch for seamless integration with the existing mmdetection3d framework.
\\
\\
\textbf{Data Preparation.}
The primary input consists of LiDAR point clouds, which are processed by the CenterPoint and FocalFormer3D detectors to generate 3D bounding boxes and associated confidence scores. Each detection contains various attributes such as position $(x, y, z) \in \mathbb{R}^3$, shape $(w,l,h) \in \mathbb{R}^3$, yaw angle ($\theta$), object class and confidence score. When training the smoother, we also perturb boxes by adding Gaussian noise (with mean zero) to positions, shape and orientations as these form the basis for further refinement.
\\

\section{Additional Results on nuScenes Dataset}
\label{sec:B}
We provide additional results on the performance of individual classes in the nuScenes test set in Table \ref{tab:class_wise}. These results were generated using a temporal window of 15 and the online variant of LiDAR MOT-DETR. We see significant improvement in all classes compared to the underlying centerpoint~\cite{Centerpoint} tracks except the trailer class. This shows how well using the temporal features in transformers improves tracking. 

\begin{table}[!htb]
  \centering
  \scriptsize
  \setlength{\tabcolsep}{1.0pt}
  \begin{center}
    \begin{tabular}{l|cccc}
      \hline
      class        & aMOTA$\uparrow$ & aMOTP$\downarrow$  & IDS$\downarrow$ & FRAG$\downarrow$ \\
      \hline
   
      bicycle          & 0.618 (\textit{0.321})     & 0.423  &  6  &      7 \\
      bus          &   0.884 (\textit{0.711})   &  0.490   &   6   &   10 \\
      car          & 0.859 (\textit{0.829})   & 0.329   &  157   &    189 \\
      motorcycle        & 0.793 (\textit{0.591})   & 0.416  &   5   &       6  \\
      pedestrain          & 0.828 (\textit{0.767})  & 0.275   &  148   &   84   \\
      trailer        & 0.561 (\textit{0.651})    & 0.875  &   3  & 17    \\
      truck          & 0.671 (\textit{0.599})      & 0.554   &  17 &  41    \\
      
    \hline
    \end{tabular}
    \begin{tabular}{cc}
    \end{tabular}
  \end{center}

  \caption{Class-wise results on nuScenes val set. Values in bracket represent CenterPoint baseline tracker.}
  \label{tab:class_wise}
\end{table}

\section{Runtime}
\label{sec:C}
The inference speed is computed using A100 GPU. During inference the smoother runs at 5.6 FPS and at 4.3 FPS for the tracker.

\section{Qualitative Results on nuScenes Validation Set}
\label{sec:D}
\begin{figure}
\begin{tabular}{ccccc}
1 &
\bmvaHangBox{\fbox{\includegraphics[width=2.15cm]{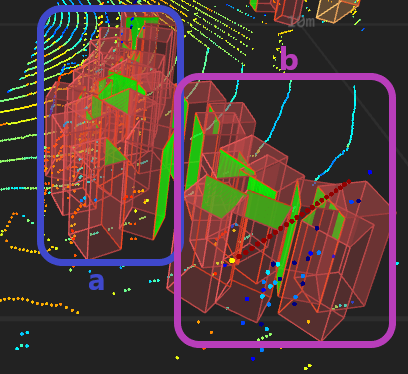}}}&
\bmvaHangBox{\fbox{\includegraphics[width=2.15cm]{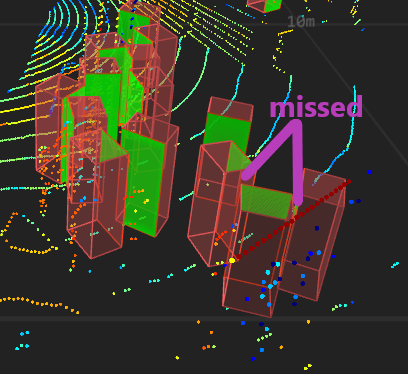}}}&
\bmvaHangBox{\fbox{\includegraphics[width=2.15cm]{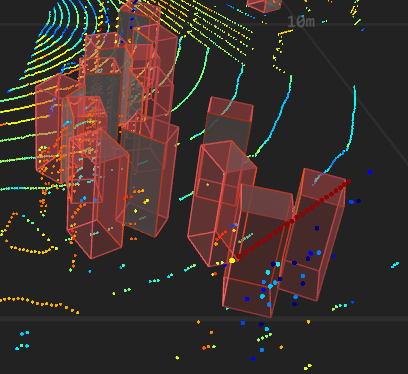}}}&
\bmvaHangBox{\fbox{\includegraphics[width=2.15cm]{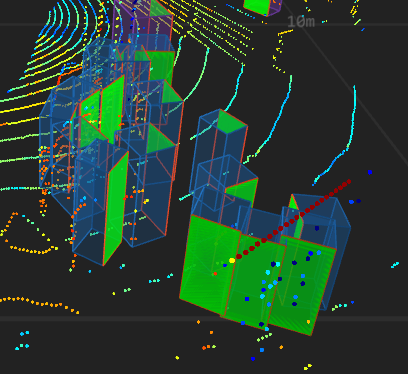}}}\\

2&
\bmvaHangBox{\fbox{\includegraphics[width=2.15cm]{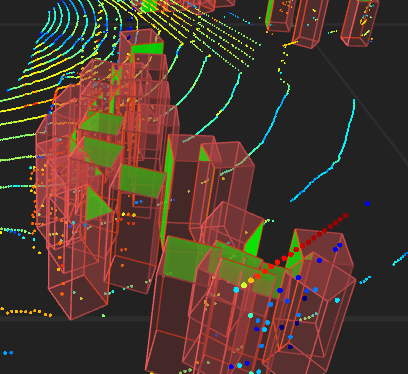}}}&
\bmvaHangBox{\fbox{\includegraphics[width=2.15cm]{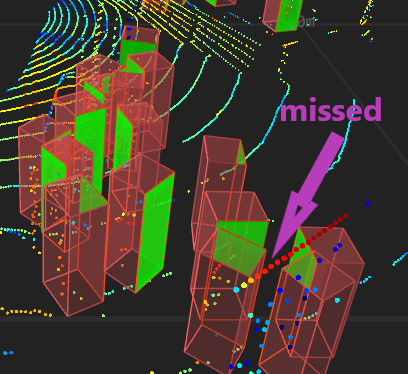}}}&
\bmvaHangBox{\fbox{\includegraphics[width=2.15cm]{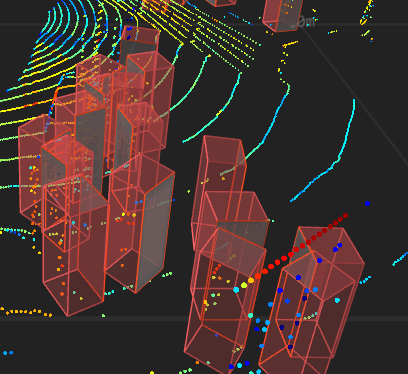}}}&
\bmvaHangBox{\fbox{\includegraphics[width=2.15cm]{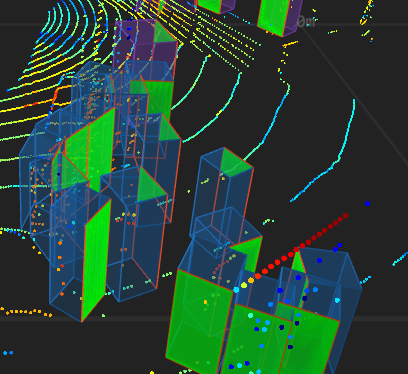}}}
\\
3&
\bmvaHangBox{\fbox{\includegraphics[width=2.15cm]{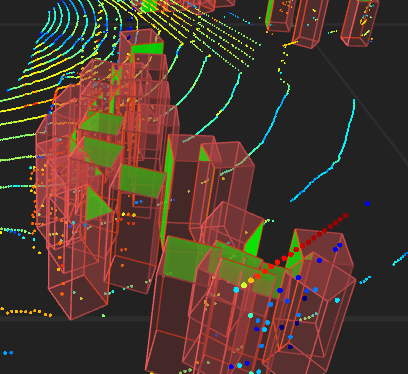}}}&
\bmvaHangBox{\fbox{\includegraphics[width=2.15cm]{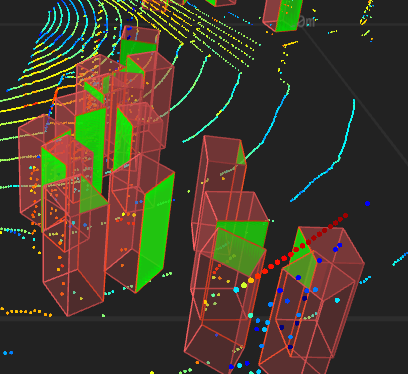}}}&
\bmvaHangBox{\fbox{\includegraphics[width=2.15cm]{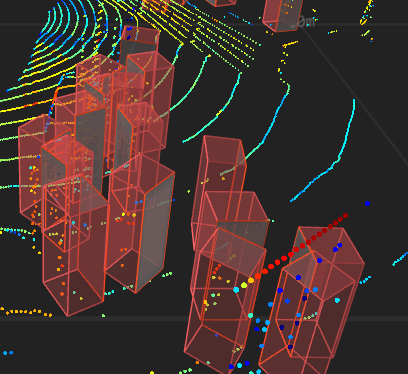}}}&
\bmvaHangBox{\fbox{\includegraphics[width=2.15cm]{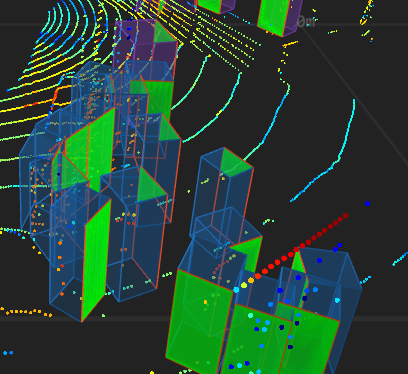}}}
\\
Frame & (detections)&(smoother)&(tracker)&(groundtruth)
\end{tabular}
\caption{Dense scene from nuScenes validation set. We recommend zooming in for better visualizations}
\label{fig:demo2}
\end{figure}

We present some dense scenarios where the LiDAR MOT-DETR works to clean up detector output as seen in figure \ref{fig:demo2}. In frame 1, we demonstrate how a crowd is sparsely detected by the CenterPoint object detector. In the area highlighted in blue rectangle and marked as \(\mathit{a}\), he smoother is able to precisely extract objects of relevance in the scene. However, in the area highlighted in purple, \(\mathit{b}\), we show some failure cases of missed objects which could be improved upon. 

We also provide a full scene demonstration from the nuscenes validation dataset in the supplementary video (scene \textbf{c525507ee2ef4c6d8bb64b0e0cf0dd32}). In this video, we comapare our work to 3DMOTFormer using Centerpoint detections in both cases. 

\end{document}